\documentclass[11pt,a4paper]{article}
\usepackage[hyperref]{eacl2021}
\usepackage{times}
\usepackage{latexsym}

\usepackage{tikz}
\usepackage{booktabs}
\usepackage{subcaption}
\usepackage{fancyvrb}
\usepackage{todonotes}
\usepackage{changes}

\usepackage{microtype}

\aclfinalcopy % Uncomment this line for the final submission
\def\aclpaperid{30} %  Enter the acl Paper ID here

\newcommand{\name}[0]{\textsc{MaChAmp}}

\title{
      \begin{minipage}[c]{\linewidth}
        \centering
        Massive Choice, Ample Tasks (\name):\\
        \begin{minipage}{1.5cm}\vspace{.0cm}    \centering\scalebox{-1}[1]{\includegraphics[width=1cm]{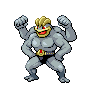}}\end{minipage}
        A Toolkit for Multi-task Learning in NLP
        \begin{minipage}{1.5cm}\vspace{.0cm}\centering \includegraphics[width=1cm]{imgs/machamp}\end{minipage} 
    \end{minipage}

}
\author{
	\begin{tabular}{ c c c c}
	\textbf{Rob van der Goot}\includegraphics[width=.4cm]{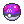}
	& \textbf{Ahmet \"{U}st\"{u}n}\includegraphics[width=.4cm]{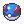} 
	& \textbf{Alan Ramponi}\includegraphics[width=.4cm]{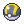}\includegraphics[width=.4cm]{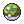}
	& \textbf{Ibrahim Sharaf}\includegraphics[width=.4cm]{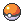}\\
	\multicolumn{4}{c}{\textbf{Barbara Plank}\includegraphics[width=.4cm]{imgs/Bag_Master_Ball_Sprite.png}}\\
	\end{tabular} \\
	\begin{tabular}{c c c}
	  \includegraphics[width=.4cm]{imgs/Bag_Master_Ball_Sprite.png}IT University of Copenhagen &
	  \includegraphics[width=.4cm]{imgs/Bag_Great_Ball_Sprite.png}University of Groningen &
	  \includegraphics[width=.4cm]{imgs/Bag_Ultra_Ball_Sprite.png}University of Trento\\ 
	\end{tabular} \\
	\begin{tabular}{c c}
      \includegraphics[width=.4cm]{imgs/Bag_Safari_Ball_Sprite.png}Fondazione the Microsoft Research - University of Trento COSBI & 
      \includegraphics[width=.4cm]{imgs/Bag_Poke_Ball_Sprite.png}Factmata \\
	\end{tabular} \\
	\begin{tabular}{c}
	\texttt{robv@itu.dk, a.ustun@rug.nl, alan.ramponi@unitn.it} \\
	\texttt{ibrahim.sharaf@factmata.com, bapl@itu.dk} \\
	\end{tabular}
}

\date{}

\begin{document}
\maketitle
\begin{abstract}
Transfer learning, particularly approaches that combine multi-task learning
with pre-trained contextualized embeddings and fine-tuning, have advanced the
field of Natural Language Processing tremendously in recent years.  In this
paper we present \name, a toolkit for easy fine-tuning of contextualized
embeddings in multi-task settings.  The benefits of \name{} are its flexible
configuration options, and the support of a variety of natural language
processing tasks in a uniform toolkit, from text classification and sequence
labeling to dependency parsing, masked language modeling, and text
generation.\footnote{The code is available at:
\url{https://github.com/machamp-nlp/machamp} (v0.2), and an instructional video
at \url{https://www.youtube.com/watch?v=DauTEdMhUDI}.}
\end{abstract}

\section{Introduction}
Multi-task learning (MTL)~\cite{caruana1993,caruana1997} has developed into a
standard repertoire in natural language processing (NLP). It enables neural
networks to learn tasks in parallel~\cite{caruana1993} while leveraging the
benefits of sharing parameters. The shift---or
``tsunami''~\cite{manning2015computational}---of deep learning in NLP has
facilitated the wide-spread use of MTL since the seminal work
by~\newcite{collobert2011natural}, which has led to a multi-task learning
``wave''~\cite{ruder-plank-2018-strong} in NLP. It has since been applied to a
wide range of NLP tasks, developing into a viable alternative to classical
pipeline approaches.  This includes early adoption in Recurrent Neural Network
models,
e.g.~\cite{lazaridou-etal-2015-combining,chrupala-etal-2015-learning,plank-etal-2016,sogaard-goldberg-2016-deep,hashimoto-etal-2017-joint},
to the use of large pre-trained language models with multi-task
objectives~\cite{radford2019language,devlin-etal-2019-bert}. MTL comes in many
flavors, based on the type of sharing, the weighting of losses, and the design
and relations of tasks and layers. In general though, outperforming single-task
settings remains a
challenge~\cite{martinez-alonso-plank-2017-multitask,clark-etal-2019-bam}. For
an overview of MTL in NLP we refer to~\newcite{ruder2017overview}.

As a separate line of research, the idea of language model pre-training and
contextual
embeddings~\cite{howard-ruder-2018-universal,peters-etal-2018-deep,devlin-etal-2019-bert}
is to pre-train rich representation on large quantities of monolingual or
multilingual text data. Taking these representations as a starting point has
led to enormous improvements across a wide variety of  NLP problems. Related to
MTL, recent research effort focuses on fine-tuning contextualized embeddings on
a variety of tasks with supervised
objectives~\cite{kondratyuk-straka-2019-75,sanh2019hierarchical,hu2020xtreme}.

We introduce \name{}, a flexible toolkit for multi-task learning and
fine-tuning of NLP problems. The main advantages of \name{} are:
\begin{itemize}
    \item Ease of configuration, especially for dealing with multiple datasets
and multi-task setups;
    \item Support of a wide range of NLP tasks, including a variety of sequence
labeling approaches, text classification, dependency parsing, masked language
modeling, and text generation (e.g., machine translation);
    \item Support of the initialization and fine-tuning of any contextualized
embeddings from Hugging Face~\cite{Wolf2019HuggingFacesTS}.
\end{itemize}

As a result, the flexibility of \name{} supports up-to-date, general-purpose
NLP (see Section~\ref{sec:task_types}).  The backbone of \name{} is
AllenNLP~\cite{gardner-etal-2018-allennlp}, a PyTorch-based \cite{pytorch}
Python library containing modules for a variety of deep learning methods and
NLP tasks. It is designed to be modular, high-level and flexible. It should be
noted that contemporary to \name{},
\texttt{jiant}~\cite{pruksachatkun-etal-2020-jiant} was developed, and AllenNLP
included multi-task learning as well since release 2.0. \name{} distinguishes
from the other toolkits by supporting simple configurations, and a variety of
multi-task settings.

\section{Model}
\label{sec:model}
In this section we will discuss the model, its supported tasks, and possible
configuration settings.

\begin{figure}
\centering
 \resizebox{6.5cm}{!}{
    \def\wordpieces#1#2#3#4#5#6#7
{
\begin{scope}
\newcount\wordPiece
\wordPiece=0
\loop
    \node (#7\the\wordPiece) [draw, minimum width=2*#4cm, minimum height=#4cm, fill=#6] at (#1 + \the\wordPiece * 4 * #4, #2) {};
    \advance \wordPiece +1
\ifnum \wordPiece<#3
\repeat
\end{scope}
}

\def\arrows#1#2#3#4#5#6
%#5 is actually not necessary, when the arrows dont have direction
{
\begin{scope}
\newcount\wordCounterOne
\wordCounterOne=0
\loop
    % I tried to make a loop in a loop, but apparently thats not posisble
    \draw [-, opacity=#6] (#1 + \wordCounterOne * 4 * #3, #2) -- (#1 + 0 * 4 * #3, #2+#5);
    \draw [-, opacity=#6] (#1 + \wordCounterOne * 4 * #3, #2) -- (#1 + 1 * 4 * #3, #2+#5);
    \draw [-, opacity=#6] (#1 + \wordCounterOne * 4 * #3, #2) -- (#1 + 2 * 4 * #3, #2+#5);
    \draw [-, opacity=#6] (#1 + \wordCounterOne * 4 * #3, #2) -- (#1 + 3 * 4 * #3, #2+#5);
    \draw [-, opacity=#6] (#1 + \wordCounterOne * 4 * #3, #2) -- (#1 + 4 * 4 * #3, #2+#5);
    %\draw [-, opacity=#6] (#1 + \wordCounterOne * 4 * #3, #2) -- (#1 + 5 * 4 * #3, #2+#5);
    \advance \wordCounterOne +1
\ifnum \wordCounterOne<#4
\repeat
\end{scope}
}

\definecolor{armygreen}{rgb}{0.29, 0.33, 0.13}
\definecolor{brickred}{rgb}{0.8, 0.25, 0.33}
\definecolor{darksalmon}{rgb}{0.91, 0.59, 0.48}
\definecolor{deeppeach}{rgb}{1.0, 0.8, 0.64}
\definecolor{deepchampagne}{rgb}{0.98, 0.84, 0.65}
\definecolor{darkgreen}{rgb}{0.0, 0.2, 0.13}
\definecolor{airforceblue}{rgb}{0.36, 0.54, 0.66}
\begin{tikzpicture}
    \path[use as bounding box] (0.25,.6) rectangle (6.65,8);

\wordpieces{0}{5.5cm}{5}{.375}{1}{brickred}{wordEnc}
\wordpieces{0}{2.5cm}{5}{.375}{1}{brickred}{wordEnc}
\node (word1) [minimum height=.25cm, text height=.25cm,text depth=.25cm] at (0,  1) {\large $<$CLS$>$};
\node (word2) [minimum height=.25cm, text height=.25cm,text depth=.25cm] at (1.5,1) {\large Smell};
%\node (word3) [minimum height=.25cm, text height=.25cm,text depth=.25cm] at (3,  1) {\Large \#\#ell};
\node (word3) [minimum height=.25cm, text height=.25cm,text depth=.25cm] at (3,1) {\large ya};
\node (word4) [minimum height=.25cm, text height=.25cm,text depth=.25cm] at (4.5,  1) {\large later};
\node (word5) [minimum height=.25cm, text height=.25cm,text depth=.25cm] at (6,1) {\large !};
%\node (inputSent) [minimum height=.25cm, text height=.25cm,text depth=.25cm] at (3,  0) {\Large Smell ya later!};
\arrows{0}{2.7}{.375}{5}{.5}{1}
\arrows{0}{5.3}{.375}{5}{-.5}{1}

\arrows{0}{3.35}{.375}{5}{.5}{.2}
\arrows{0}{4.65}{.375}{5}{-.5}{.2}

\node (Bert) [rectangle, fill,line width=.05cm, draw, opacity=.5, deepchampagne, minimum width=7cm, minimum height=1.5cm] at (3,4) {};
\node (Bert) [rectangle, line width=.05cm, minimum width=7cm, minimum height=1.5cm] at (3,4) {Contextualized Embeddings};

\node (feats1) [rectangle, line width=.05cm, darkgreen, minimum width=.85cm, minimum height=3.5cm] at (0,4) {};
\node (feats2) [rectangle, line width=.05cm, darkgreen, minimum width=.85cm, minimum height=3.5cm] at (1.5,4) {};
\node (feats3) [rectangle, line width=.05cm, darkgreen, minimum width=.85cm, minimum height=3.5cm] at (3,4) {};
\node (feats4) [rectangle, line width=.05cm, darkgreen, minimum width=.85cm, minimum height=3.5cm] at (4.5,4) {};
\node (feats5) [rectangle, line width=.05cm, darkgreen, minimum width=.85cm, minimum height=3.5cm] at (6,4) {};

\node (sentOut) [minimum height=.25cm, text height=.25cm,text depth=.25cm] at (0,7.5) {negative};
\node (upos1) [minimum height=.25cm, text height=.25cm,text depth=.25cm] at (1.5,7.5) {VERB};
\node (upos2) [minimum height=.25cm, text height=.25cm,text depth=.25cm] at (3,7.5) {PRON};
\node (upos3) [minimum height=.25cm, text height=.25cm,text depth=.25cm] at (4.5, 7.5) {ADV};
\node (upos4) [minimum height=.25cm, text height=.25cm,text depth=.25cm] at (6,7.5) {PUNCT};

\draw [->] (feats1) -- ([yshift=.25cm]sentOut.south);
\draw [->] (feats2) -- ([yshift=.25cm]upos1.south);
\draw [->] (feats3) -- ([yshift=.25cm]upos2.south);
\draw [->] (feats4) -- ([yshift=.25cm]upos3.south);
\draw [->] (feats5) -- ([yshift=.25cm]upos4.south);

\draw [->] (word1) -- (feats1);
\draw [->] (word2) -- (feats2);
\draw [->] (word3) -- (feats3);
\draw [->] (word4) -- (feats4);
\draw [->] (word5) -- (feats5);

%\draw [->] (2.75, 0.4) -- (2.75,  .9);

\node (polDecoder) [rectangle, line width=.05cm, draw, fill=airforceblue, text width=1.45cm, opacity=.8, minimum width=1.45cm, minimum height=1cm] at (0.35,6.65) {Sentiment Decoder};
\node (uposDecoder) [rectangle, line width=.05cm, draw, fill=airforceblue, opacity=.8, minimum width=5.1cm, minimum height=1cm] at (3.9,6.65) {UPOS Decoder};

%\node (layerAttent) [rectangle, line width=.05cm, draw, fill=darkgreen, opacity=.2, minimum width=8cm, minimum height=.35cm] at (3.75,6.25) {};
%\node (layerAttent) [rectangle, line width=.05cm, draw, color=darkgreen, minimum width=8cm, minimum height=.35cm] at (3.75,6.25) {};
%\node (layerAttent) [rectangle, line width=.05cm, minimum width=8cm, minimum height=.35cm] at (3.75,6.25) {\small Layer $\;\;\;$ Attention};

%Alternative:  "gotta catch em' all!"
%Alternative: "I choose you, Machamp!"
\end{tikzpicture}
    }
    \caption{Overview of \name{}, when training jointly for sentiment analysis
and POS tagging. A shared encoding representation and task-specific decoders
are exploited to accomplish both tasks.}
    \label{fig:overview}
\end{figure}
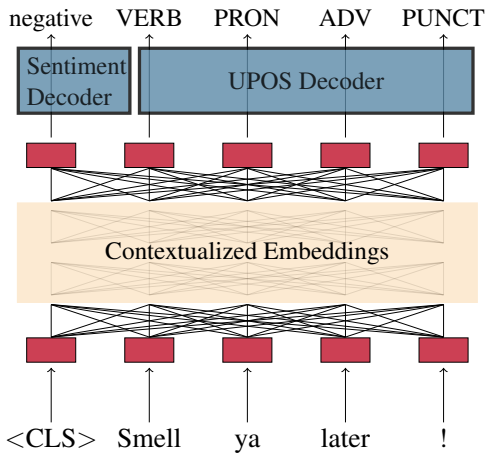

\subsection{Model overview}
An overview of the model is shown in Figure~\ref{fig:overview}. \name~takes a
pre-trained contextualized model as initial encoder, and fine-tunes its layers
by applying an inverse square root learning rate decay with linear warm-up
\cite{howard-ruder-2018-universal}, according to a given set of downstream
tasks. For the task-specific predictions, each task has its own decoder, which
is trained for the corresponding task. The model defaults to the
embedding-specific tokenizer in Hugging
Face~\cite{Wolf2019HuggingFacesTS}.\footnote{This includes both the
pre-tokenization (in the traditional sense) and the subword segmentation.}

When multiple datasets are used for training, they are first separately split
into batches so that each batch only contains instances from one dataset.
Batches are then concatenated and shuffled before training.  This means that
small datasets will be underrepresented, which can be overcome by smoothing the
dataset sampling (Section~\ref{sec:hyperconf}).  During decoding, the loss
function is only activated for tasks which are present in the current batch.
By default, all tasks have an equal weight in the loss function. The loss
weight can be tuned (Section~\ref{sec:datasetConfig}).

\subsection{Supported task types}
\label{sec:task_types}
We here describe the tasks \name{} supports.

\paragraph{\textsc{seq}} For traditional token-level sequence prediction tasks,
like part-of-speech tagging. \name{} uses greedy decoding with a softmax output
layer on the output of the contextual embeddings.

\paragraph{\textsc{string2string}} An extension to \textsc{seq}, which learns a
conversion for each input token to its label.  Instead of predicting the labels
directly, the model can now learn to predict the conversion. This strategy is
commonly used for lemmatization~\cite{chrupala2006,kondratyuk-straka-2019-75},
where it greatly reduces the label vocabulary. We use the transformation
algorithm from UDPipe-Future~\cite{straka-2018-udpipe}, which was also used
by~\newcite{kondratyuk-straka-2019-75}. 

\paragraph{\textsc{seq\_bio}} A variant of \textsc{seq} which exploits
conditional random fields~\cite{lafferty2001crf} as decoder, masked to enforce
outputs following the BIO tagging scheme.

\paragraph{\textsc{multiseq}} An extension to \textsc{seq} which supports the
prediction of multiple labels per token. Specifically, for some sequence
labeling tasks it is unknown beforehand how many labels each token should get.
We compute a probability score for each label, employing binary cross-entropy
as loss, and outputting all the labels that exceed a certain threshold. The
threshold can be set in the dataset configuration file.

\paragraph{\textsc{dependency}} For dependency parsing, \name{} uses the deep
biaffine parser~\cite{dozat2016deep} as implemented by
AllenNLP~\cite{gardner-etal-2018-allennlp}, with the Chu-Liu/Edmonds
algorithm~\cite{chu1965shortest,edmonds1967optimum} for decoding the tree.

\paragraph{\textsc{mlm}} For masked language modeling, our implementation
follows the original BERT settings~\cite{devlin-etal-2019-bert}. The chance
that a token is masked is 15\%, of which 80\% are masked with a \texttt{[MASK]}
token, 10\% with a random token, and 10\% are left unchanged. We do not include
the next sentence prediction task following~\newcite{liu2019roberta}, for
simplicity and efficiency.  We use a cross entropy loss, and the language model
heads from the defined Hugging Face embeddings~\cite{Wolf2019HuggingFacesTS}.
It assumes raw text files as input, so no \texttt{column\_idx} has to be
defined (See Section \ref{sec:data_format}).

\paragraph{\textsc{classification}} For text classification, it predicts a
label for every text instance by using the embedding of the first token, which
is commonly a special token (e.g.\ \texttt{[CLS]} or \texttt{<s>}). For tasks
which model a relation between multiple sentences (e.g., textual entailment), a
special token (e.g.\ \texttt{[SEP]}) is automatically inserted between the
sentences to inform the model about the sentence boundaries.

\paragraph{\textsc{seq2seq}} For text generation, \name{} employs the sequence
to sequence (encoder-decoder) paradigm~\cite{sutskever2014sequence}. We use a
recurrent neural network decoder, which suits the auto-regressive nature of the
machine translation tasks~\cite{cho-etal-2014-learning} and an attention
mechanism to avoid compressing the whole source sentence into a fixed-length
vector~\cite{bahdanau2015neural}.

\section{Usage}
\begin{figure}
\begin{subfigure}{\columnwidth}
\begin{Verbatim}[frame=single]
smell   VERB
ya  PRON
later   ADV
!   PUNCT
\end{Verbatim}
\caption{Example of a token-level file format (e.g., for POS tagging), where
words are in column \texttt{word\_idx}=0, and a single layer of corresponding
annotations is in column \texttt{column\_idx}=1.}
\label{fig:ex:seq}
\end{subfigure}

\begin{subfigure}{\columnwidth}
\begin{Verbatim}

\end{Verbatim}
\begin{Verbatim}[frame=single]
smell ya later !    negative
\end{Verbatim}
\caption{Example of a sentence-level file format (e.g., for sentiment
classification), where only a sentence is required and is defined in column 0
(i.e., \texttt{sent\_idxs}=[0]) and a single layer of annotation is in the
second column (\texttt{column\_idx}=1).}
\label{fig:ex:clas}
\end{subfigure}
\caption{Examples of data file formats.}
\end{figure}

To use \name{}, one needs a configuration file, input data and a command to
start the training or prediction. In this section we will describe each of
these requirements.

\subsection{Data format} \label{sec:data_format}
\name{} supports two types of data formats for annotated data,\footnote{The MLM
task does not require annotation, thus a raw text file can be provided.} which
correspond to the level of annotation (Section~\ref{sec:task_types}). For
token-level tasks, we will use the term ``token-level file format'', whereas
for sentence-level task, we will use ``sentence-level file format''.

The token-level file format is similar to the tab-separated CoNLL
format~\cite{tjong-kim-sang-de-meulder-2003-introduction}. It assumes one token
per line (on a column index \texttt{word\_idx}), with each annotation layer
following each token separated by a tab character (each on a column index
\texttt{column\_idx}) (Figure~\ref{fig:ex:seq}). Token sequences (e.g.,
sentences) are delimited by an empty line. Comments are lines on top of the
sequence (which have a different number of columns with respect to "token
lines").\footnote{We do not identify comments based on lines starting with a
`\#', because datasets might have tokens that begin with `\#'.} It should be
noted that for dependency parsing, the format assumes the relation label to be
on the \texttt{column\_idx} and the head index on the following column.
Further, we also support the UD format by removing multi-word tokens and empty
nodes using the UD-conversion-tools~\cite{agic-etal-2016-multilingual}.

The sentence-level file format (used for text classification and text
generation) is similar (Figure~\ref{fig:ex:clas}), and also supports multiple
inputs having the same annotation layers. A list of one or more column indices
can be defined (i.e., \texttt{sent\_idxs}) to enable modeling the relation
between any arbitrary number of sentences.

\subsection{Configuration}
The model requires two configuration files, one that specifies the datasets and
tasks, and one for the hyperparameters. For the hyperparameters, a default
option is provided (\texttt{configs/params.json}, see
Section~\ref{sec:tuning}).

\begin{figure}
\fontsize{8}{9.6}
\begin{Verbatim}[frame=single]
{ "UD": {
    "train_data_path": "data/ewt.train",
    "validation_data_path": "data/ewt.dev",
    "word_idx": 1,
    "tasks": {
      "lemma": {
        "task_type": "string2string",
        "column_idx": 2
      },
      "upos": {
        "task_type": "seq",
        "column_idx": 3
  } } }
  "RTE": {
    "train_data_path": "data/RTE.train",
    "validation_data_path": "data/RTE.dev",
    "sent_idxs": [0,1],
    "tasks": {
      "rte": {
        "task_type": "classification",
        "column_idx": 2
} } } }
\end{Verbatim}
\caption{Example dataset configuration file to predict UPOS, lemmas, and
textual entailment simultaneously.}
\label{fig:config}
\end{figure}

\subsubsection{Dataset configuration}
\label{sec:dataconf}

An example of a dataset configuration file is shown in Figure~\ref{fig:config}.
On the first level, the dataset names are specified (i.e., ``UD'' and ``RTE''),
which should be unique identifiers. Each of these datasets needs at least a
\texttt{train\_data\_path}, a \texttt{validation\_data\_path}, a
\texttt{word\_idx} or \texttt{sent\_idxs}, and a list of \texttt{tasks}
(corresponding to the layers of annotation, see Section~\ref{sec:data_format}).

For each of the defined tasks, the user is required to define the
\texttt{task\_type} (Section~\ref{sec:task_types}), and the column index from
which to read the relevant labels (i.e., \texttt{column\_idx}). On top of this
template, the following options can be passed on the task level:

\label{sec:datasetConfig}

\paragraph{Metric} For each task type, a commonly used metric is set as default
metric. However, one can override the default by specifying a different metric
at the task level. Supported metrics are `acc', `las', `micro-f1', `macro-f1',
`span\_f1', 'multi\_span\_f1', 'bleu' and 'perplexity'.

\paragraph{Loss weight} In multi-task settings, not all tasks might be equally
important, or some tasks might just be harder to learn, and therefore should
gain more weight during training. This can be tuned by setting the
\texttt{loss\_weight} parameter on the task level (by default the value is 1.0
for all tasks).

\paragraph{Dataset embedding}
\newcite{ammar-etal-2016-many} have shown that embedding which language an
instance belongs to can be beneficial for multilingual models. Later
work~\cite{stymne-etal-2018-parser,wagner-etal-2020-treebank} has also shown
that more fine-grained distinctions on the dataset level\footnote{These are
called treebank embeddings in their work. We will use the more general term
``dataset embeddings'', which would often roughly correspond to languages
and/or domains/genres.} can be beneficial when training on multiple datasets
within the same language (family). In previous work, this embedding is usually
concatenated to the word embedding before the encoding. However, in
contextualized embeddings, the word embeddings themselves are commonly used as
encoder, hence we concatenate the dataset embeddings in between the encoder and
the decoder. This parameter is set on the dataset level with
\texttt{dataset\_embed\_idx}, which specifies the column to read the dataset ID
from. Setting \texttt{dataset\_embed\_idx} to -1 will use the dataset name as
specified in the json file as ID.

\paragraph{Max sentences} In order to limit the maximum number of sentences
that are used during training, \texttt{max\_sents} is used. This is done before
the sampling smoothing (Section~\ref{sec:hyperconf}), if both are enabled. It
should be noted that the specified number will be taken from the top of the
dataset.

\subsubsection{Hyperparameter configuration} \label{sec:hyperconf} Whereas most
of the hyperparameters can simply be changed from the default configuration
provided in \texttt{configs/params.json}, we would like to highlight two main
settings.

\paragraph{Pre-trained embeddings} The name/path to pre-trained Hugging Face
embeddings\footnote{\url{https://huggingface.co/models}} can be set in the
configuration file at the  \texttt{transformer\_model} key;
\texttt{transformer\_dim} might be adapted accordingly to reflect the
embeddings dimension.

\begin{figure}
\includegraphics[width=\columnwidth]{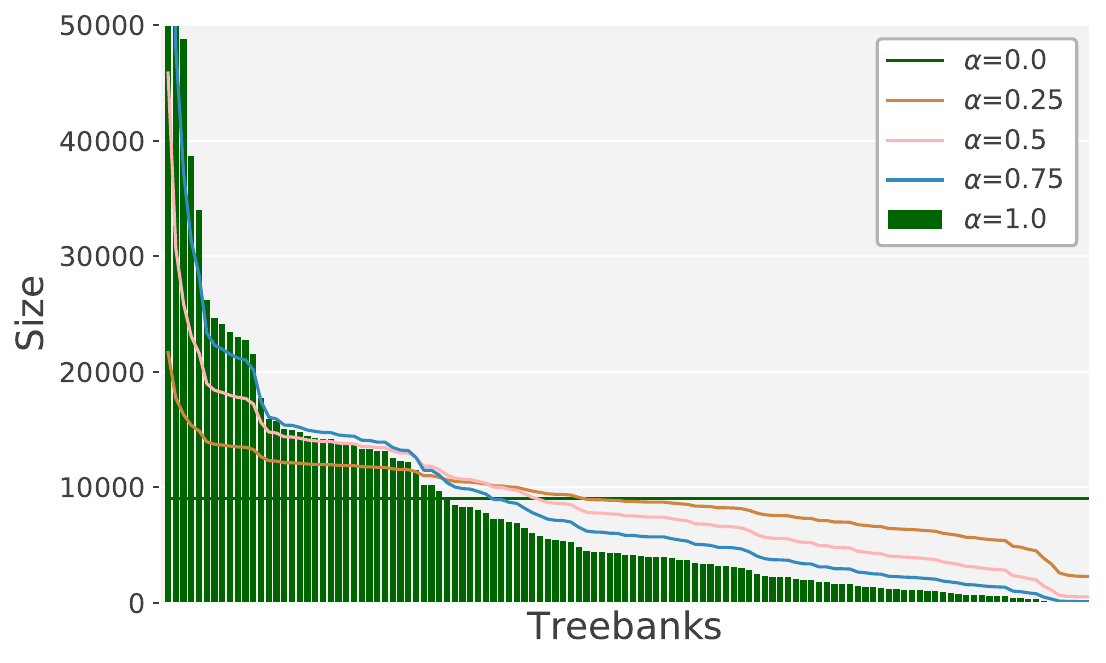}
\caption{Effect of the sampling parameter $\alpha$ on the training sets of Universal Dependencies 2.6 data.}
\label{fig:sampling}
\end{figure}

\paragraph{Dataset sampling}
To avoid larger datasets from overwhelming the model, \name{} can re-sample
multiple datasets according to a multinomial distribution, similar as used
by~\newcite{conneau2019cross}. \name{} performs the sampling on the batch
level, and shuffles after each epoch (so it can see a larger variety of
instances for downsampled datasets). The formula is:

\begin{equation}
 \lambda = \frac{1}{p_i} * \frac{p_i^\alpha }{\sum_i p_i^ \alpha}
\end{equation}

\noindent where $p_i$ is the probability that a random sample is from dataset
$i$, and $\alpha$ is a hyperparameter that can be set.  Setting $\alpha$=1.0
means using the default sizes, and $\alpha$=0.0 results in one average amount
of batches for each dataset, similar to~\newcite{sanh2019hierarchical}. The
effect of different settings of $\alpha$ for the Universal Dependencies 2.6
data is shown in Figure~\ref{fig:sampling}. Smoothing can be enabled in the
hyperparameters configuration file at the \texttt{sampling\_smoothing} key.

\subsection{Training}

Given the setup illustrated in the previous sections, a model can be trained
via the following command. It assumes the configuration
(Figure~\ref{fig:config}) is saved in \texttt{configs/upos-lemma-rte.json}.
\vspace{.15cm}

\small
\begin{verbatim}
python3 train.py --dataset_config \
 configs/upos-lemma-rte.json 
\end{verbatim}
\normalsize
\vspace{.15cm}

By default, the model and the logs will be written to
\texttt{logs/$<$JSONNAME$>$/$<$DATE$>$}. The name of the directory can be set
manually by providing \texttt{--name $<$NAME$>$}. Further, \texttt{--device
<ID>} can be used to specify which GPU to use, otherwise the CPU will be used.
As a default, \texttt{train.py} uses \texttt{configs/params.json} for the
hyperparameters, but this can be overridden by using
\texttt{--parameters\_config <CONFIG FILE>}.

\begin{table}
\hspace{-.3cm}
\resizebox{1.05\columnwidth}{!}{
  \begin{tabular}{l r r r}
    \toprule
    Parameter & Value & Range \\
    \midrule
    Optimizer                   & Adam & \\
    $\beta_1$, $\beta_2$        & 0.9, 0.99 & \\
    Dropout                     & 0.2& 0.1, 0.2, 0.3\\
    Epochs                      & 20 & \\
    Batch size                  & 32 & \\
    Learning rate (LR)          & 1e-4 & 1e-3, 1e-4, 1e-5\\
    LR scheduler                & slanted triangular & \\
    Weight decay                & 0.01 & \\
    Decay factor                & 0.38 & .35, .38, .5\\
    Cut fraction                & 0.2 & .1, .2, .3 \\
\bottomrule
  \end{tabular}
  }
  \caption{Final parameter settings, incl.\ tested ranges.}
  \label{tab:params}
\end{table}

\subsection{Inference}

Prediction can be done with:
\vspace{.15cm}

\small
\begin{verbatim}
python3 predict.py \
 logs/<NAME>/<DATE>/model.tar.gz \
 <INPUT FILE> <OUTPUT FILE>
\end{verbatim}
\normalsize
\vspace{.15cm}

It requires the path to the best model (serialized during training) stored as
\texttt{model.tar.gz} in the logs directory as specified above. By default, the
data is assumed to be in the same format as the training data (i.e., with the
same number of \texttt{column\_idx} columns), but \texttt{--raw\_text} can be
specified to read a data file containing raw texts with one sentence per line.
For models trained on multiple datasets (as ``UD'' and ``RTE'' in
Figure~\ref{fig:config}), \texttt{--dataset <NAME>} can be used to specify
which dataset to use in order to predict all tasks within that dataset.

\begin{table}
\small
\centering
\begin{tabular}{l r r}
\toprule
Task & Reference & \name{} \\
\midrule
\textbf{EWT2.2} & \color{darkblue}{Kondratyuk et al. (2019)}  \\
~~~~UPOS$^*$ & 96.82 & \textbf{97.07}\\
~~~~Lemma$^*$ &  97.97 & \textbf{98.14}\\
~~~~Feats$^*$ &  97.27 & \textbf{97.41}\\
~~~~LAS$^*$ & 89.38 & \textbf{89.80} \\
\midrule
\textbf{GLUE} & \newcite{devlin-etal-2019-bert}\\
~~~~CoLA &  \textbf{60.5}& 53.7\\
~~~~MNLI &  \textbf{86.7} & 83.9\\
~~~~MNLI-mis & \textbf{85.9} & 82.7\\
~~~~MRPC & \textbf{89.3} & 87.2 \\
~~~~QNLI &  \textbf{92.7} & 90.8\\
~~~~QQP & \textbf{72.1} & 69.1 \\
~~~~RTE &  \textbf{70.1} & 60.0\\
~~~~SST-2 &  \textbf{94.9} & 92.5\\
\midrule
\textbf{WMT14} & \newcite{liu2020deep} \\
~~~~EN-DE & \textbf{30.1} & 24.7 \\
\textbf{IWSLT15} & \newcite{zaheer2018adaptive} & \\
~~~~EN-VI & \textbf{29.27} & 24.72 \\
\bottomrule
\end{tabular}
\caption{Scores of single task models on test data for three popular datasets
and a variety of tasks. $^*$one joint model. For the GLUE data, BERT-large
(English) and tokenized BLEU are used for fair comparison.}
\label{tab:single}
\end{table}

\section{Hyperparameter Tuning}
\label{sec:tuning}
In this section we describe the procedure how we determined robust default
parameters for \name{}; note that the goal is not to beat the state-of-the-art,
but to reach competitive performance for multiple tasks
simultaneously.\footnote{ Compared to \name{} v0.1~\cite{van2020massive} we
removed parameters with negligible effects (word dropout, layer dropout,
adaptive softmax, and layer attention).}

For the tuning of hyperparameters, we used the GLUE classification
datasets~\cite{wang-etal-2018-glue,warstadt2019neural,socher-etal-2013-recursive,dolan-brockett-2005-automatically,cer-etal-2017-semeval,N18-1101,rajpurkar-etal-2018-know,bentivogli2009fifth,levesque2012winograd}
and the English Web Treebank (EWT 2.6)~\cite{silveira-etal-2014-gold} with
multilingual
BERT\footnote{\url{https://github.com/google-research/bert/blob/master/multilingual.md}}
(mBERT) as embeddings.\footnote{We capped the dataset sizes to a maximum of
20,000 sentences for efficiency reasons.} For each of these setups, we averaged
the scores over all datasets/tasks and perform a grid search. The best
hyperparameters across all datasets are reported in Table~\ref{tab:params} and
are the defaults values for \name{}.

\section{Evaluation}

\subsection{Single task evaluation}
As a starting point, we evaluate single task models to ensure our
implementations are competitive with the state-of-the-art. We report scores on
dependency parsing (EWT), the GLUE classification tasks, and machine
translation (WMT14 DE-EN~\cite{bojar-etal-2014-findings}, IWSLT15
EN-VI~\cite{cettolo2014iwslt}) using mBERT as our embeddings.\footnote{For the
sake of comparison we use BERT-large for GLUE, and EWT version 2.2.}
Table~\ref{tab:single} reports our results on the test sets compared to
previous work. For all UD tasks, we score slightly higher, whereas for GLUE
tasks we score consistently lower compared to the references. This is mostly
due to differences in fine-tuning strategies, as implementations themselves are
highly similar. Scores on the machine translation tasks show the largest drops,
indicating that task-specific fine-tuning and pre-processing might be
necessary.

\begin{table}
    \centering
    \resizebox{.9\columnwidth}{!}{

    \begin{tabular}{l r r}
        \toprule
        Setup           & UD (LAS)& GLUE (Acc)\\
        \midrule
        Single          & 72.22 & \textbf{82.38} \\
        All             & 72.82 & 80.96 \\
        Smoothed        & \textbf{73.74} & 81.87 \\
        Dataset embed.$^*$  & 72.76 & --\\
        Sep. decoder$^*$    & 73.69 & --\\
        \bottomrule
    \end{tabular}}
    \caption{Average results over all development sets. Dataset embeddings and
a separate decoder have not been tested in GLUE, because each dataset is
annotated for a different task. $^*$includes dataset smoothing.}
    \label{tab:eval}
\end{table}

\subsection{Multi-dataset evaluation}
We evaluate the effect of a variety of multi-dataset settings on all GLUE and
UD treebanks (v2.7) on the test splits. It should be noted that the UD
treebanks all have the same tasks, as opposed to GLUE.  First, we jointly train
on all datasets (\textsc{All}), then we attempt to improve performance on
smaller sets by enabling the sampling smoothing (\textsc{Smoothed},
Section~\ref{sec:hyperconf}, we set $\alpha=0.5$). Furthermore, we attempt to
improve the performance by informing the decoder of the dataset through dataset
embeddings (\textsc{Dataset embed.}, Section~\ref{sec:datasetConfig}) or by
giving each dataset its own decoder (\textsc{Sep. decoder}).  Results
(Table~\ref{tab:eval}) show that multi-task learning is only beneficial for
performance when training on the same set of tasks (i.e., UD), dataset
smoothing is helpful, dataset embeddings and separate decoders do not improve
upon smoothing on average.

\begin{table}
\centering
    \resizebox{.9\columnwidth}{!}{
\begin{tabular}{l r r r r r r r}
\toprule
Model\textbackslash Size & 0 & $<$1k & $<$10k & $>$10k   \\
\midrule
Single & 43.5 & 15.1 & 57.9 & 80.1 \\
All & 44.5 & 37.1 & 66.4 & 80.3 \\
Smoothed & 44.3 & \textbf{45.4} & 67.1 & 80.3 \\
Dataset embed.$^*$ & 43.9 & 36.5 & \textbf{67.8} & \textbf{81.0} \\
Sep. decoder$^*$ & \textbf{45.1} & 37.7 & 66.5 & 80.9 \\
\bottomrule
\end{tabular}}
\caption{Average LAS scores on test splits of UD treebanks grouped by training
size (in number of sentences). $^*$includes dataset smoothing.}
\label{tab:analysisUD}
\end{table}

For analysis purposes, we group the UD treebanks based on training size, and
also evaluate UD treebanks which have no training split (zero-shot).  For the
zero-shot experiments, we select a proxy parser based on word overlap of the
first 10 sentences of the target test data and the source training
data.\footnote{Scores on individual sets and proxy treebanks can be found in
the Appendix.} Results on the UD data (Table~\ref{tab:analysisUD}) show that
multi-task learning is mostly beneficial for medium-sized datasets ($<$1k and
$<$10k). For these datasets, the combination of smoothing and dataset
embeddings are the most promising settings. Perhaps surprisingly, the zero-shot
datasets ($<$1k) have a higher LAS as compared to the small datasets and using
a separate decoder based on the proxy treebank is the best setting; this is
mainly because for many small datasets there is no other in-language training
treebank.  For the GLUE tasks (Table~\ref{tab:analysisGlue}, Appendix),
multi-task learning is only beneficial for the RTE data. This is to be
expected, as the tasks are different in this setup, and training data is
generally larger. Dataset smoothing here prevents the model from dropping too
much in performance, as it outperforms \textsc{All} for 7 out of 9 tasks.

\section{Conclusion}
We introduced \name{}, a powerful toolkit for multi-task learning supporting a wide range of NLP tasks. We also provide initial experiments demonstrating the usefulness of some of its options. We learned that multi-task learning is mostly beneficial for setups in which multiple datasets are annotated for the same set of tasks, 
and that dataset embeddings can still be useful when employing contextualized embeddings. However, the current experiments are just scratching the surface of \name{}'s capabilities, as a wide variety of tasks and multi-task settings is supported. 

\section*{Acknowledgments}
\noindent We would like to thank Anouck Braggaar, Max M\"{u}ller-Eberstein and Kristian Nørgaard Jensen for testing development versions. Furthermore, we thank Rik van Noord for his participation in the video, and providing an early use-case for \name{}~\cite{van-noord-etal-2020-character}.
This research was supported by an Amazon Research Award, an STSM in the Multi3Generation COST action (CA18231), a visit supported by COSBI, grant 9063-00077B (Danmarks
Frie Forskningsfond), and Nvidia corporation for sponsoring Titan GPUs. We thank the NLPL laboratory and the HPC team at ITU for the computational resources used in this work.

\bibliography{papers, datasets}

\begin{thebibliography}{205}
\expandafter\ifx\csname natexlab\endcsname\relax\def\natexlab#1{#1}\fi

\bibitem[{Abeill{\'e} et~al.(2000)Abeill{\'e}, Cl{\'e}ment, and
  Kinyon}]{abeille-etal-2000-building}
Anne Abeill{\'e}, Lionel Cl{\'e}ment, and Alexandra Kinyon. 2000.
\newblock \href {http://www.lrec-conf.org/proceedings/lrec2000/pdf/230.pdf}
  {Building a treebank for {F}rench}.
\newblock In \emph{Proceedings of the Second International Conference on
  Language Resources and Evaluation ({LREC}{'}00)}, Athens, Greece. European
  Language Resources Association (ELRA).

\bibitem[{Aepli and Clematide(2018)}]{aepli2018parsing}
No\"emi Aepli and Simon Clematide. 2018.
\newblock {Parsing approaches for Swiss German}.
\newblock In \emph{{Proceedings of the 3rd Swiss Text Analytics Conference
  (SwissText), Winterthur, Switzerland}}.

\bibitem[{Agi{\'c} et~al.(2016)Agi{\'c}, Johannsen, Plank,
  Mart{\'\i}nez~Alonso, Schluter, and S{\o}gaard}]{agic-etal-2016-multilingual}
{\v{Z}}eljko Agi{\'c}, Anders Johannsen, Barbara Plank, H{\'e}ctor
  Mart{\'\i}nez~Alonso, Natalie Schluter, and Anders S{\o}gaard. 2016.
\newblock \href {https://doi.org/10.1162/tacl_a_00100} {Multilingual projection
  for parsing truly low-resource languages}.
\newblock \emph{Transactions of the Association for Computational Linguistics},
  4:301--312.

\bibitem[{Agi{\'c} and Ljube{\v{s}}i{\'c}(2015)}]{agic-ljubesic-2015-universal}
{\v{Z}}eljko Agi{\'c} and Nikola Ljube{\v{s}}i{\'c}. 2015.
\newblock \href {https://www.aclweb.org/anthology/W15-5301} {{U}niversal
  {D}ependencies for {C}roatian (that work for {S}erbian, too)}.
\newblock In \emph{The 5th Workshop on {B}alto-{S}lavic Natural Language
  Processing}, pages 1--8, Hissar, Bulgaria. INCOMA Ltd. Shoumen, BULGARIA.

\bibitem[{Ahrenberg(2015)}]{ahrenberg-2015-converting}
Lars Ahrenberg. 2015.
\newblock \href {https://www.aclweb.org/anthology/W15-2103} {Converting an
  {E}nglish-{S}wedish parallel treebank to {U}niversal {D}ependencies}.
\newblock In \emph{Proceedings of the Third International Conference on
  Dependency Linguistics (Depling 2015)}, pages 10--19, Uppsala, Sweden.
  Uppsala University, Uppsala, Sweden.

\bibitem[{Alfieri and Tamburini(2016)}]{alfieri2016almost}
Linda Alfieri and Fabio Tamburini. 2016.
\newblock (almost) automatic conversion of the {Venice Italian Treebank} into
  the merged {Italian Dependency Treebank} format.
\newblock In \emph{CLiC-it/EVALITA}.

\bibitem[{Alfina et~al.(2020)Alfina, Budi, and Suhartanto}]{alfina2020tree}
Ika Alfina, Indra Budi, and Heru Suhartanto. 2020.
\newblock Tree rotations for dependency trees: Converting the
  head-directionality of noun phrases.
\newblock \emph{Journal of Computer Science}, 16(11):1585--1597.

\bibitem[{Alonso and Zeman(2016)}]{alonso2016universal}
H{\'e}ctor~Mart{\'\i}nez Alonso and Daniel Zeman. 2016.
\newblock Universal {D}ependencies for the {AnCora} treebanks.
\newblock \emph{Procesamiento del Lenguaje Natural}, 57:91--98.

\bibitem[{Ammar et~al.(2016)Ammar, Mulcaire, Ballesteros, Dyer, and
  Smith}]{ammar-etal-2016-many}
Waleed Ammar, George Mulcaire, Miguel Ballesteros, Chris Dyer, and Noah~A.
  Smith. 2016.
\newblock \href {https://doi.org/10.1162/tacl_a_00109} {Many languages, one
  parser}.
\newblock \emph{Transactions of the Association for Computational Linguistics},
  4:431--444.

\bibitem[{Aquino et~al.(2020)Aquino, de~Leon, and Bacolod}]{UD_Tagalog-Ugnayan}
Angelina Aquino, Franz de~Leon, and Mary~Ann Bacolod. 2020.
\newblock {UD\_Tagalog-Ugnayan}.
\newblock \url{https://github.com/UniversalDependencies/UD_Tagalog-Ugnayan}.

\bibitem[{Aragon(2018)}]{aragon2018variaccoes}
Carolina~Coelho Aragon. 2018.
\newblock Varia{\c{c}}{\~o}es estil{\'i}sticas e sociais no discurso dos
  falantes akunts{\'u}.
\newblock \emph{Polifonia}, 25(38.1):90--103.

\bibitem[{Aranzabe et~al.(2015)Aranzabe, Atutxa, Bengoetxea, de~Ilarraza,
  Goenaga, Gojenola, and Uria}]{aranzabe2015automatic}
Maria~Jesus Aranzabe, Aitziber Atutxa, Kepa Bengoetxea, Arantza~Diaz
  de~Ilarraza, Iakes Goenaga, Koldo Gojenola, and Larraitz Uria. 2015.
\newblock Automatic conversion of the {B}asque dependency treebank to universal
  dependencies.
\newblock In \emph{Proceedings of the fourteenth international workshop on
  treebanks an linguistic theories (TLT14)}, pages 233--241.

\bibitem[{Asahara et~al.(2018)Asahara, Kanayama, Tanaka, Miyao, Uematsu, Mori,
  Matsumoto, Omura, and Murawaki}]{asahara-etal-2018-universal}
Masayuki Asahara, Hiroshi Kanayama, Takaaki Tanaka, Yusuke Miyao, Sumire
  Uematsu, Shinsuke Mori, Yuji Matsumoto, Mai Omura, and Yugo Murawaki. 2018.
\newblock \href {https://www.aclweb.org/anthology/L18-1287} {{U}niversal
  {D}ependencies version 2 for {J}apanese}.
\newblock In \emph{Proceedings of the Eleventh International Conference on
  Language Resources and Evaluation ({LREC} 2018)}, Miyazaki, Japan. European
  Language Resources Association (ELRA).

\bibitem[{Badmaeva and Tyers(2017)}]{badmaeva:2017}
Elena Badmaeva and Francis~M. Tyers. 2017.
\newblock Dependency treebank for {B}uryat.
\newblock In \emph{Proceedings of the 15th International Workshop on Treebanks
  and Linguistic Theories (TLT15)}, pages 1--12.

\bibitem[{Bahdanau et~al.(2015)Bahdanau, Cho, and Bengio}]{bahdanau2015neural}
Dzmitry Bahdanau, Kyunghyun Cho, and Yoshua Bengio. 2015.
\newblock \href {http://arxiv.org/abs/1409.0473} {Neural machine translation by
  jointly learning to align and translate}.
\newblock In \emph{3rd International Conference on Learning Representations,
  ICLR 2015, Conference Track Proceedings}, San Diego, CA, USA.

\bibitem[{Bamman and Crane(2011)}]{bamman2011ancient}
David Bamman and Gregory Crane. 2011.
\newblock The ancient {G}reek and {L}atin dependency treebanks.
\newblock In \emph{Language technology for cultural heritage}, pages 79--98.
  Springer.

\bibitem[{Barbu~Mititelu et~al.(2016)Barbu~Mititelu, Ion, Simionescu, Irimia,
  and Perez}]{barbu2016romanian}
Verginica Barbu~Mititelu, Radu Ion, Radu Simionescu, Elena Irimia, and
  Cenel-Augusto Perez. 2016.
\newblock The {R}omanian treebank annotated according to {U}niversal
  {D}ependencies.
\newblock In \emph{Proceedings of the tenth international conference on natural
  language processing (hrtal2016)}.

\bibitem[{Batchelor(2019)}]{batchelor-2019-universal}
Colin Batchelor. 2019.
\newblock \href {https://www.aclweb.org/anthology/W19-6902} {Universal
  dependencies for {S}cottish {G}aelic: syntax}.
\newblock In \emph{Proceedings of the Celtic Language Technology Workshop},
  pages 7--15, Dublin, Ireland. European Association for Machine Translation.

\bibitem[{Behzad and Zeldes(2020)}]{behzad-zeldes-2020-cross}
Shabnam Behzad and Amir Zeldes. 2020.
\newblock \href {https://www.aclweb.org/anthology/2020.wac-1.7} {A cross-genre
  ensemble approach to robust {R}eddit part of speech tagging}.
\newblock In \emph{Proceedings of the 12th Web as Corpus Workshop}, pages
  50--56, Marseille, France. European Language Resources Association.

\bibitem[{Bej{\v{c}}ek et~al.(2013)Bej{\v{c}}ek, Haji{\v{c}}ov{\'a},
  Haji{\v{c}}, J{\'\i}nov{\'a}, Kettnerov{\'a}, Kol{\'a}{\v{r}}ov{\'a},
  Mikulov{\'a}, M{\'\i}rovsk{\`y}, Nedoluzhko, Panevov{\'a}
  et~al.}]{bejvcek2013prague}
Eduard Bej{\v{c}}ek, Eva Haji{\v{c}}ov{\'a}, Jan Haji{\v{c}}, Pavl{\'\i}na
  J{\'\i}nov{\'a}, V{\'a}clava Kettnerov{\'a}, Veronika Kol{\'a}{\v{r}}ov{\'a},
  Marie Mikulov{\'a}, Ji{\v{r}}{\'\i} M{\'\i}rovsk{\`y}, Anna Nedoluzhko,
  Jarmila Panevov{\'a}, et~al. 2013.
\newblock Prague dependency treebank 3.0.

\bibitem[{Bentivogli et~al.(2009)Bentivogli, Dagan, Dang, Giampiccolo, and
  Magnini}]{bentivogli2009fifth}
Luisa Bentivogli, Ido Dagan, Hoa~Trang Dang, Danilo Giampiccolo, and Bernardo
  Magnini. 2009.
\newblock \href
  {https://tac.nist.gov/publications/2009/additional.papers/RTE5\_overview.proceedings.pdf}
  {The fifth {PASCAL} recognizing textual entailment challenge.}
\newblock In \emph{Proceedings of the Second Text Analysis Conference, {TAC}
  2009}, Gaithersburg, Maryland, USA.

\bibitem[{Berzak et~al.(2016)Berzak, Kenney, Spadine, Wang, Lam, Mori, Garza,
  and Katz}]{berzak-etal-2016-universal}
Yevgeni Berzak, Jessica Kenney, Carolyn Spadine, Jing~Xian Wang, Lucia Lam,
  Keiko~Sophie Mori, Sebastian Garza, and Boris Katz. 2016.
\newblock \href {https://doi.org/10.18653/v1/P16-1070} {{U}niversal
  {D}ependencies for learner {E}nglish}.
\newblock In \emph{Proceedings of the 54th Annual Meeting of the Association
  for Computational Linguistics (Volume 1: Long Papers)}, pages 737--746,
  Berlin, Germany. Association for Computational Linguistics.

\bibitem[{Bhat et~al.(2018)Bhat, Bhat, Shrivastava, and
  Sharma}]{bhat-etal-2018-universal}
Irshad Bhat, Riyaz~A. Bhat, Manish Shrivastava, and Dipti Sharma. 2018.
\newblock \href {https://doi.org/10.18653/v1/N18-1090} {{U}niversal
  {D}ependency parsing for {H}indi-{E}nglish code-switching}.
\newblock In \emph{Proceedings of the 2018 Conference of the North {A}merican
  Chapter of the Association for Computational Linguistics: Human Language
  Technologies, Volume 1 (Long Papers)}, pages 987--998, New Orleans,
  Louisiana. Association for Computational Linguistics.

\bibitem[{Bhat et~al.(2016)Bhat, Bhatt, Farudi, Klassen, Narasimhan, Palmer,
  Rambow, Sharma, Vaidya, Vishnu et~al.}]{bhathindi}
Riyaz~Ahmad Bhat, Rajesh Bhatt, Annahita Farudi, Prescott Klassen, Bhuvana
  Narasimhan, Martha Palmer, Owen Rambow, Dipti~Misra Sharma, Ashwini Vaidya,
  Sri~Ramagurumurthy Vishnu, et~al. 2016.
\newblock The hindi/urdu treebank project.
\newblock In \emph{Handbook of Linguistic Annotation}. Springer Press.

\bibitem[{Bielinskiene et~al.(2016)Bielinskiene, Boizou, and
  Kovalevskaite}]{bielinskiene2016lithuanian}
Agne Bielinskiene, Loic Boizou, and Jolanta Kovalevskaite. 2016.
\newblock Lithuanian dependency treebank.
\newblock In \emph{Human Language Technologies--The Baltic Perspective:
  Proceedings of the Seventh International Conference Baltic HLT 2016}, volume
  289, page 107. IOS Press.

\bibitem[{Blodgett et~al.(2018)Blodgett, Wei, and
  O{'}Connor}]{blodgett-etal-2018-twitter}
Su~Lin Blodgett, Johnny Wei, and Brendan O{'}Connor. 2018.
\newblock \href {https://doi.org/10.18653/v1/P18-1131} {{T}witter {U}niversal
  {D}ependency parsing for {A}frican-{A}merican and mainstream {A}merican
  {E}nglish}.
\newblock In \emph{Proceedings of the 56th Annual Meeting of the Association
  for Computational Linguistics (Volume 1: Long Papers)}, pages 1415--1425,
  Melbourne, Australia. Association for Computational Linguistics.

\bibitem[{Bojar et~al.(2014)Bojar, Buck, Federmann, Haddow, Koehn, Leveling,
  Monz, Pecina, Post, Saint-Amand, Soricut, Specia, and
  Tamchyna}]{bojar-etal-2014-findings}
Ond{\v{r}}ej Bojar, Christian Buck, Christian Federmann, Barry Haddow, Philipp
  Koehn, Johannes Leveling, Christof Monz, Pavel Pecina, Matt Post, Herve
  Saint-Amand, Radu Soricut, Lucia Specia, and Ale{\v{s}} Tamchyna. 2014.
\newblock \href {https://doi.org/10.3115/v1/W14-3302} {Findings of the 2014
  workshop on statistical machine translation}.
\newblock In \emph{Proceedings of the Ninth Workshop on Statistical Machine
  Translation}, pages 12--58, Baltimore, Maryland, USA. Association for
  Computational Linguistics.

\bibitem[{Bonfante et~al.(2018)Bonfante, Guillaume, and
  Perrier}]{bonfante2018application}
Guillaume Bonfante, Bruno Guillaume, and Guy Perrier. 2018.
\newblock \emph{Application of Graph Rewriting to Natural Language Processing}.
\newblock Wiley Online Library.

\bibitem[{Borges~V{\"o}lker et~al.(2019)Borges~V{\"o}lker, Wendt, Hennig, and
  K{\"o}hn}]{borges-volker-etal-2019-hdt}
Emanuel Borges~V{\"o}lker, Maximilian Wendt, Felix Hennig, and Arne K{\"o}hn.
  2019.
\newblock \href {https://doi.org/10.18653/v1/W19-8006} {{HDT}-{UD}: A very
  large {U}niversal {D}ependencies treebank for {G}erman}.
\newblock In \emph{Proceedings of the Third Workshop on Universal Dependencies
  (UDW, SyntaxFest 2019)}, pages 46--57, Paris, France. Association for
  Computational Linguistics.

\bibitem[{Bosco et~al.(2014)Bosco, Dell'Orletta, Montemagni, Sanguinetti, and
  Simi}]{bosco2014evalita}
Cristina Bosco, Felice Dell'Orletta, Simonetta Montemagni, Manuela Sanguinetti,
  and Maria Simi. 2014.
\newblock The {EVALITA} 2014 dependency parsing task.
\newblock In \emph{EVALITA 2014 Evaluation of NLP and Speech Tools for
  Italian}, pages 1--8. Pisa University Press.

\bibitem[{Bouma and van Noord(2017)}]{bouma-van-noord-2017-increasing}
Gosse Bouma and Gertjan van Noord. 2017.
\newblock \href {https://www.aclweb.org/anthology/W17-0403} {Increasing return
  on annotation investment: The automatic construction of a {U}niversal
  {D}ependency treebank for {D}utch}.
\newblock In \emph{Proceedings of the {N}o{D}a{L}i{D}a 2017 Workshop on
  Universal Dependencies ({UDW} 2017)}, pages 19--26, Gothenburg, Sweden.
  Association for Computational Linguistics.

\bibitem[{Braggaar and van~der Goot(2021)}]{braggaar2021}
Anouck Braggaar and Rob van~der Goot. 2021.
\newblock Challenges in annotating and parsing spoken, code-switched,
  {Frisian-Dutch} data.
\newblock In \emph{Proceedings of the Second Workshop on Domain Adaptation for
  Natural Language Processing}.

\bibitem[{Brants et~al.(2004)Brants, Dipper, Eisenberg, Hansen-Schirra,
  K{\"o}nig, Lezius, Rohrer, Smith, and Uszkoreit}]{brants2004tiger}
Sabine Brants, Stefanie Dipper, Peter Eisenberg, Silvia Hansen-Schirra, Esther
  K{\"o}nig, Wolfgang Lezius, Christian Rohrer, George Smith, and Hans
  Uszkoreit. 2004.
\newblock {TIGER}: Linguistic interpretation of a german corpus.
\newblock \emph{Research on language and computation}, 2(4):597--620.

\bibitem[{Caron et~al.(2019)Caron, Courtin, Gerdes, and
  Kahane}]{caron-etal-2019-surface}
Bernard Caron, Marine Courtin, Kim Gerdes, and Sylvain Kahane. 2019.
\newblock \href {https://doi.org/10.18653/v1/W19-7803} {A surface-syntactic
  {UD} treebank for {N}aija}.
\newblock In \emph{Proceedings of the 18th International Workshop on Treebanks
  and Linguistic Theories (TLT, SyntaxFest 2019)}, pages 13--24, Paris, France.
  Association for Computational Linguistics.

\bibitem[{Caruana(1993)}]{caruana1993}
Rich Caruana. 1993.
\newblock \href {https://doi.org/10.1016/b978-1-55860-307-3.50012-5}
  {{Multitask learning: A knowledge-based source of inductive bias}}.
\newblock In \emph{Proceedings of the Tenth International Conference on Machine
  Learning}, pages 41--48, Amherst, MA, USA.

\bibitem[{Caruana(1997)}]{caruana1997}
Rich Caruana. 1997.
\newblock \href {https://doi.org/10.1007/978-1-4615-5529-2\_5} {Multitask
  learning}.
\newblock In \emph{Learning to learn}, pages 95--133. Springer.

\bibitem[{Cecchini et~al.(2018)Cecchini, Passarotti, Marongiu, and
  Zeman}]{cecchini-etal-2018-challenges}
Flavio~Massimiliano Cecchini, Marco Passarotti, Paola Marongiu, and Daniel
  Zeman. 2018.
\newblock \href {https://doi.org/10.18653/v1/W18-6004} {Challenges in
  converting the index {T}homisticus treebank into {U}niversal {D}ependencies}.
\newblock In \emph{Proceedings of the Second Workshop on Universal Dependencies
  ({UDW} 2018)}, pages 27--36, Brussels, Belgium. Association for Computational
  Linguistics.

\bibitem[{{\v{C}}{\'e}pl{\"o}(2018)}]{vceplo2018constituent}
Slavom{\'\i}r {\v{C}}{\'e}pl{\"o}. 2018.
\newblock Constituent order in {M}altese: A quantitative analysis.

\bibitem[{Cer et~al.(2017)Cer, Diab, Agirre, Lopez-Gazpio, and
  Specia}]{cer-etal-2017-semeval}
Daniel Cer, Mona Diab, Eneko Agirre, I{\~n}igo Lopez-Gazpio, and Lucia Specia.
  2017.
\newblock \href {https://doi.org/10.18653/v1/S17-2001} {{S}em{E}val-2017 task
  1: Semantic textual similarity multilingual and crosslingual focused
  evaluation}.
\newblock In \emph{Proceedings of the 11th International Workshop on Semantic
  Evaluation ({S}em{E}val-2017)}, pages 1--14, Vancouver, Canada. Association
  for Computational Linguistics.

\bibitem[{{\c{C}}etino{\u{g}}lu and
  {\c{C}}{\"o}ltekin(2019)}]{cetinoglu-coltekin-2019-challenges}
{\"O}zlem {\c{C}}etino{\u{g}}lu and {\c{C}}a{\u{g}}r{\i} {\c{C}}{\"o}ltekin.
  2019.
\newblock \href {https://doi.org/10.18653/v1/W19-7809} {Challenges of
  annotating a code-switching treebank}.
\newblock In \emph{Proceedings of the 18th International Workshop on Treebanks
  and Linguistic Theories (TLT, SyntaxFest 2019)}, pages 82--90, Paris, France.
  Association for Computational Linguistics.

\bibitem[{Cettolo et~al.(2014)Cettolo, Jan, Sebastian, Bentivogli, Cattoni, and
  Federico}]{cettolo2014iwslt}
Mauro Cettolo, Niehues Jan, St{\"u}ker Sebastian, Luisa Bentivogli, Roldano
  Cattoni, and Marcello Federico. 2014.
\newblock The {IWSLT} 2014 evaluation campaign.
\newblock In \emph{International Workshop on Spoken Language Translation}, Lake
  Tahoe, CA, USA.

\bibitem[{Cho et~al.(2014)Cho, van Merri{\"e}nboer, Gulcehre, Bahdanau,
  Bougares, Schwenk, and Bengio}]{cho-etal-2014-learning}
Kyunghyun Cho, Bart van Merri{\"e}nboer, Caglar Gulcehre, Dzmitry Bahdanau,
  Fethi Bougares, Holger Schwenk, and Yoshua Bengio. 2014.
\newblock \href {https://doi.org/10.3115/v1/D14-1179} {Learning phrase
  representations using {RNN} encoder{--}decoder for statistical machine
  translation}.
\newblock In \emph{Proceedings of the 2014 Conference on Empirical Methods in
  Natural Language Processing ({EMNLP})}, pages 1724--1734, Doha, Qatar.
  Association for Computational Linguistics.

\bibitem[{Chrupa{\l}a et~al.(2015)Chrupa{\l}a, K{\'a}d{\'a}r, and
  Alishahi}]{chrupala-etal-2015-learning}
Grzegorz Chrupa{\l}a, {\'A}kos K{\'a}d{\'a}r, and Afra Alishahi. 2015.
\newblock \href {https://doi.org/10.3115/v1/P15-2019} {Learning language
  through pictures}.
\newblock In \emph{Proceedings of the 53rd Annual Meeting of the Association
  for Computational Linguistics and the 7th International Joint Conference on
  Natural Language Processing (Volume 2: Short Papers)}, pages 112--118,
  Beijing, China. Association for Computational Linguistics.

\bibitem[{Chrupała(2006)}]{chrupala2006}
Grzegorz Chrupała. 2006.
\newblock \href
  {http://journal.sepln.org/sepln/ojs/ojs/index.php/pln/article/viewFile/2741/1259}
  {Simple data-driven context-sensitive lemmatization}.
\newblock \emph{SEPLN}, 37:121--127.

\bibitem[{Chu(1965)}]{chu1965shortest}
Yoeng-Jin Chu. 1965.
\newblock On the shortest arborescence of a directed graph.
\newblock \emph{Scientia Sinica}, 14:1396--1400.

\bibitem[{Chun et~al.(2018)Chun, Han, Hwang, and
  Choi}]{chun-etal-2018-building}
Jayeol Chun, Na-Rae Han, Jena~D. Hwang, and Jinho~D. Choi. 2018.
\newblock \href {https://www.aclweb.org/anthology/L18-1347} {Building
  {U}niversal {D}ependency treebanks in {K}orean}.
\newblock In \emph{Proceedings of the Eleventh International Conference on
  Language Resources and Evaluation ({LREC} 2018)}, Miyazaki, Japan. European
  Language Resources Association (ELRA).

\bibitem[{Cignarella et~al.(2019)Cignarella, Bosco, and
  Rosso}]{cignarella-etal-2019-presenting}
Alessandra~Teresa Cignarella, Cristina Bosco, and Paolo Rosso. 2019.
\newblock \href {https://doi.org/10.18653/v1/W19-7723} {Presenting
  {TWITTIR{\`O}}-{UD}: An {I}talian {T}witter treebank in {U}niversal
  {D}ependencies}.
\newblock In \emph{Proceedings of the Fifth International Conference on
  Dependency Linguistics (Depling, SyntaxFest 2019)}, pages 190--197, Paris,
  France. Association for Computational Linguistics.

\bibitem[{Clark et~al.(2019)Clark, Luong, Khandelwal, Manning, and
  Le}]{clark-etal-2019-bam}
Kevin Clark, Minh-Thang Luong, Urvashi Khandelwal, Christopher~D. Manning, and
  Quoc~V. Le. 2019.
\newblock \href {https://doi.org/10.18653/v1/P19-1595} {{BAM}! born-again
  multi-task networks for natural language understanding}.
\newblock In \emph{Proceedings of the 57th Annual Meeting of the Association
  for Computational Linguistics}, pages 5931--5937, Florence, Italy.
  Association for Computational Linguistics.

\bibitem[{Collobert et~al.(2011)Collobert, Weston, Bottou, Karlen, Kavukcuoglu,
  and Kuksa}]{collobert2011natural}
Ronan Collobert, Jason Weston, L{\'e}on Bottou, Michael Karlen, Koray
  Kavukcuoglu, and Pavel Kuksa. 2011.
\newblock \href {http://dl.acm.org/citation.cfm?id=2078186} {Natural language
  processing (almost) from scratch}.
\newblock \emph{Journal of Machine Learning Research}, 12:2493--2537.

\bibitem[{C{\"o}ltekin(2015)}]{coltekin2015grammar}
Cagr{\i} C{\"o}ltekin. 2015.
\newblock A grammar-book treebank of {T}urkish.
\newblock In \emph{Proceedings of the 14th workshop on Treebanks and Linguistic
  Theories (TLT 14)}, pages 35--49.

\bibitem[{Conneau and Lample(2019)}]{conneau2019cross}
Alexis Conneau and Guillaume Lample. 2019.
\newblock \href
  {https://proceedings.neurips.cc/paper/2019/hash/c04c19c2c2474dbf5f7ac4372c5b9af1-Abstract.html}
  {Cross-lingual language model pretraining}.
\newblock In \emph{Advances in Neural Information Processing Systems}, pages
  7059--7069, Vancouver, Canada.

\bibitem[{Davidson et~al.(2019)Davidson, Yu, and
  Yu}]{davidson-etal-2019-dependency}
Sam Davidson, Dian Yu, and Zhou Yu. 2019.
\newblock \href {https://doi.org/10.18653/v1/D19-1162} {Dependency parsing for
  spoken dialog systems}.
\newblock In \emph{Proceedings of the 2019 Conference on Empirical Methods in
  Natural Language Processing and the 9th International Joint Conference on
  Natural Language Processing (EMNLP-IJCNLP)}, pages 1513--1519, Hong Kong,
  China. Association for Computational Linguistics.

\bibitem[{Derin(2020)}]{ud_old_turkish_tonqq_2020}
Mehmet~Oguz Derin. 2020.
\newblock {UD\_Old\_Turkish-Tonqq}.
\newblock \url{https://github.com/UniversalDependencies/UD_Old_Turkish-Tonqq}.

\bibitem[{Devlin et~al.(2019)Devlin, Chang, Lee, and
  Toutanova}]{devlin-etal-2019-bert}
Jacob Devlin, Ming-Wei Chang, Kenton Lee, and Kristina Toutanova. 2019.
\newblock \href {https://doi.org/10.18653/v1/N19-1423} {{BERT}: Pre-training of
  deep bidirectional transformers for language understanding}.
\newblock In \emph{Proceedings of the 2019 Conference of the North {A}merican
  Chapter of the Association for Computational Linguistics: Human Language
  Technologies, Volume 1 (Long and Short Papers)}, pages 4171--4186,
  Minneapolis, Minnesota. Association for Computational Linguistics.

\bibitem[{Dione(2019)}]{dione-2019-developing}
Cheikh~Bamba Dione. 2019.
\newblock \href {https://doi.org/10.18653/v1/W19-8003} {Developing {U}niversal
  {D}ependencies for {W}olof}.
\newblock In \emph{Proceedings of the Third Workshop on Universal Dependencies
  (UDW, SyntaxFest 2019)}, pages 12--23, Paris, France. Association for
  Computational Linguistics.

\bibitem[{Dirix et~al.(2017)Dirix, Augustinus, van Niekerk, and
  Van~Eynde}]{dirix-etal-2017-universal}
Peter Dirix, Liesbeth Augustinus, Daniel van Niekerk, and Frank Van~Eynde.
  2017.
\newblock \href {https://www.aclweb.org/anthology/W17-0405} {{U}niversal
  {D}ependencies for {A}frikaans}.
\newblock In \emph{Proceedings of the {N}o{D}a{L}i{D}a 2017 Workshop on
  Universal Dependencies ({UDW} 2017)}, pages 38--47, Gothenburg, Sweden.
  Association for Computational Linguistics.

\bibitem[{Dobrovoljc et~al.(2017)Dobrovoljc, Erjavec, and
  Krek}]{dobrovoljc-etal-2017-universal}
Kaja Dobrovoljc, Toma{\v{z}} Erjavec, and Simon Krek. 2017.
\newblock \href {https://doi.org/10.18653/v1/W17-1406} {The {U}niversal
  {D}ependencies treebank for {S}lovenian}.
\newblock In \emph{Proceedings of the 6th Workshop on {B}alto-{S}lavic Natural
  Language Processing}, pages 33--38, Valencia, Spain. Association for
  Computational Linguistics.

\bibitem[{Dobrovoljc and Nivre(2016)}]{dobrovoljc-nivre-2016-universal}
Kaja Dobrovoljc and Joakim Nivre. 2016.
\newblock \href {https://www.aclweb.org/anthology/L16-1248} {The {U}niversal
  {D}ependencies treebank of spoken {S}lovenian}.
\newblock In \emph{Proceedings of the Tenth International Conference on
  Language Resources and Evaluation ({LREC}'16)}, pages 1566--1573,
  Portoro{\v{z}}, Slovenia. European Language Resources Association (ELRA).

\bibitem[{Dolan and Brockett(2005)}]{dolan-brockett-2005-automatically}
William~B. Dolan and Chris Brockett. 2005.
\newblock \href {https://www.aclweb.org/anthology/I05-5002} {Automatically
  constructing a corpus of sentential paraphrases}.
\newblock In \emph{Proceedings of the Third International Workshop on
  Paraphrasing ({IWP}2005)}, pages 9--16, Jeju Island, Korea.

\bibitem[{Dozat and Manning(2017)}]{dozat2016deep}
Timothy Dozat and Christopher~D Manning. 2017.
\newblock \href {https://nlp.stanford.edu/pubs/dozat2017deep.pdf} {Deep
  biaffine attention for neural dependency parsing}.
\newblock In \emph{Proceedings of 5th International Conference on Learning
  Representations, ICLR 2017, Conference Track Proceedings}, Toulon, France.

\bibitem[{Droganova et~al.(2018)Droganova, Lyashevskaya, and
  Zeman}]{droganova2018data}
Kira Droganova, Olga Lyashevskaya, and Daniel Zeman. 2018.
\newblock Data conversion and consistency of monolingual corpora: {Russian UD}
  treebanks.
\newblock In \emph{Proceedings of the 17th international workshop on treebanks
  and linguistic theories (tlt 2018)}, 155, pages 53--66.

\bibitem[{Dwivedi and Easha(2017)}]{dwivedi2017universal}
Puneet Dwivedi and Guha Easha. 2017.
\newblock Universal {D}ependencies for {S}anskrit.
\newblock \emph{International Journal of Advance Research, Ideas and
  Innovations in Technology}, 3(4).

\bibitem[{Eckhoff et~al.(2018)Eckhoff, Bech, Bouma, Eide, Haug, Haugen, and
  J{\o}hndal}]{eckhoff2018proiel}
Hanne Eckhoff, Kristin Bech, Gerlof Bouma, Kristine Eide, Dag Haug, Odd~Einar
  Haugen, and Marius J{\o}hndal. 2018.
\newblock The {PROIEL} treebank family: a standard for early attestations of
  {Indo-European} languages.
\newblock \emph{Language Resources and Evaluation}, 52(1):29--65.

\bibitem[{Eckhoff and Berdi{\v{c}}evskis(2015)}]{eckhoff2015linguistics}
Hanne~Martine Eckhoff and Aleksandrs Berdi{\v{c}}evskis. 2015.
\newblock Linguistics vs. digital editions: The {T}roms{\o} {Old Russian} and
  {OCS} treebank.
\newblock \emph{Scripta \& e-Scripta}, 14(15):9--25.

\bibitem[{Edmonds(1967)}]{edmonds1967optimum}
Jack Edmonds. 1967.
\newblock Optimum branchings.
\newblock \emph{Journal of Research of the national Bureau of Standards B},
  71(4):233--240.

\bibitem[{Eli et~al.(2016)Eli, Mushajiang, Yibulayin, Abiderexiti, and
  Liu}]{eli-etal-2016-universal}
Marhaba Eli, Weinila Mushajiang, Tuergen Yibulayin, Kahaerjiang Abiderexiti,
  and Yan Liu. 2016.
\newblock \href {https://www.aclweb.org/anthology/W16-5206} {Universal
  dependencies for {U}yghur}.
\newblock In \emph{Proceedings of the Third International Workshop on Worldwide
  Language Service Infrastructure and Second Workshop on Open Infrastructures
  and Analysis Frameworks for Human Language Technologies
  ({WLSI}/{OIAF}4{HLT}2016)}, pages 44--50, Osaka, Japan. The COLING 2016
  Organizing Committee.

\bibitem[{Freitas(2017)}]{freitas2017posse}
Mar{\'\i}lia Fernanda Pereira~de Freitas. 2017.
\newblock A posse em apurin{\~a}: Descri{\c{c}}{\~a}o de constru{\c{c}}{\~o}es
  atributivas e predicativas em compara{\c{c}}{\~a}o com outras l{\'\i}nguas
  aru{\'a}k.
\newblock \emph{Bel{\'e}m: Programa de P{\'o}s-Gradua{\c{c}}{\~a}o em Letras,
  Universidade Federal do Par{\'a} (Tese de Doutorado)}.

\bibitem[{Garcia(2016)}]{garcia2016universal}
Marcos Garcia. 2016.
\newblock Universal dependencies guidelines for the {Galician-TreeGal}
  treebank.
\newblock Technical report, Technical Report, LyS Group, Universidade da
  Coruna.

\bibitem[{Gardner et~al.(2018)Gardner, Grus, Neumann, Tafjord, Dasigi, Liu,
  Peters, Schmitz, and Zettlemoyer}]{gardner-etal-2018-allennlp}
Matt Gardner, Joel Grus, Mark Neumann, Oyvind Tafjord, Pradeep Dasigi,
  Nelson~F. Liu, Matthew Peters, Michael Schmitz, and Luke Zettlemoyer. 2018.
\newblock \href {https://doi.org/10.18653/v1/W18-2501} {{A}llen{NLP}: A deep
  semantic natural language processing platform}.
\newblock In \emph{Proceedings of Workshop for {NLP} Open Source Software
  ({NLP}-{OSS})}, pages 1--6, Melbourne, Australia. Association for
  Computational Linguistics.

\bibitem[{Gerardi(2020)}]{UD_Tupinamba-TuDeT}
Fabr\`{i}cio~Ferraz Gerardi. 2020.
\newblock {UD\_Tupinamba-TuDeT}.
\newblock \url{https://github.com/UniversalDependencies/UD_Tupinamba-TuDeT}.

\bibitem[{Gerardi(2021)}]{Tudet}
Fabr\`{i}cio~Ferraz Gerardi. 2021.
\newblock The structure of {M}undurukú.

\bibitem[{G{\"o}k{\i}rmak and Tyers(2017)}]{gokirmak-tyers-2017-dependency}
Memduh G{\"o}k{\i}rmak and Francis~M. Tyers. 2017.
\newblock \href {https://www.aclweb.org/anthology/W17-6509} {A dependency
  treebank for {K}urmanji {K}urdish}.
\newblock In \emph{Proceedings of the Fourth International Conference on
  Dependency Linguistics (Depling 2017)}, pages 64--72, Pisa,Italy.
  Link{\"o}ping University Electronic Press.

\bibitem[{G{\'o}mez~Guinovart(2017)}]{gomez2017recursos}
Xavier G{\'o}mez~Guinovart. 2017.
\newblock Recursos integrados da lingua galega para a investigaci{\'o}n
  ling{\"u}{\'\i}stica.
\newblock \emph{Gallaecia. Estudos de ling{\"u}{\'\i}stica portuguesa e galega.
  Santiago de Compostela: Universidade de Santiago}, pages 1037--1048.

\bibitem[{van~der Goot and van
  Noord(2018)}]{van-der-goot-van-noord-2018-modeling}
Rob van~der Goot and Gertjan van Noord. 2018.
\newblock \href {https://doi.org/10.18653/v1/D18-1542} {Modeling input
  uncertainty in neural network dependency parsing}.
\newblock In \emph{Proceedings of the 2018 Conference on Empirical Methods in
  Natural Language Processing}, pages 4984--4991, Brussels, Belgium.
  Association for Computational Linguistics.

\bibitem[{van~der Goot et~al.(2020)van~der Goot, {\"U}st{\"u}n, Ramponi, and
  Plank}]{van2020massive}
Rob van~der Goot, Ahmet {\"U}st{\"u}n, Alan Ramponi, and Barbara Plank. 2020.
\newblock \href {https://arxiv.org/abs/2005.14672v2} {Massive choice, ample
  tasks ({MaChAmp}): A toolkit for multi-task learning in {NLP}}.
\newblock \emph{arXiv preprint arXiv:2005.14672v2}.

\bibitem[{Gruzitis et~al.(2018)Gruzitis, Pretkalnina, Saulite, Rituma,
  Nespore-Berzkalne, Znotins, and Paikens}]{gruzitis-etal-2018-creation}
Normunds Gruzitis, Lauma Pretkalnina, Baiba Saulite, Laura Rituma, Gunta
  Nespore-Berzkalne, Arturs Znotins, and Peteris Paikens. 2018.
\newblock \href {https://www.aclweb.org/anthology/L18-1714} {Creation of a
  balanced state-of-the-art multilayer corpus for {NLU}}.
\newblock In \emph{Proceedings of the Eleventh International Conference on
  Language Resources and Evaluation ({LREC} 2018)}, Miyazaki, Japan. European
  Language Resources Association (ELRA).

\bibitem[{Guillaume et~al.(2019)Guillaume, de~Marneffe, and
  Perrier}]{guillaume2019conversion}
Bruno Guillaume, Marie-Catherine de~Marneffe, and Guy Perrier. 2019.
\newblock Conversion et am{\'e}liorations de corpus du {F}ran{\c{c}}ais
  annot{\'e}s en {U}niversal {D}ependencies.
\newblock \emph{Traitement Automatique des Langues}, 60(2):71--95.

\bibitem[{Haji{\v{c}} et~al.(2009)Haji{\v{c}}, Smr{\v{z}}, Zem{\'a}nek, Pajas,
  {\v{S}}naidauf, Be{\v{s}}ka, Kr{\'a}cmar, and Hassanov{\'a}}]{UD_Arabic-PADT}
Jan Haji{\v{c}}, Otakar Smr{\v{z}}, Petr Zem{\'a}nek, Petr Pajas, Jan
  {\v{S}}naidauf, Emanuel Be{\v{s}}ka, Jakub Kr{\'a}cmar, and Kamila
  Hassanov{\'a}. 2009.
\newblock {Prague Arabic dependency treebank 1.0}.

\bibitem[{Hashimoto et~al.(2017)Hashimoto, Xiong, Tsuruoka, and
  Socher}]{hashimoto-etal-2017-joint}
Kazuma Hashimoto, Caiming Xiong, Yoshimasa Tsuruoka, and Richard Socher. 2017.
\newblock \href {https://doi.org/10.18653/v1/D17-1206} {A joint many-task
  model: Growing a neural network for multiple {NLP} tasks}.
\newblock In \emph{Proceedings of the 2017 Conference on Empirical Methods in
  Natural Language Processing}, pages 1923--1933, Copenhagen, Denmark.
  Association for Computational Linguistics.

\bibitem[{Haug and J{\o}hndal(2008)}]{haug2008creating}
Dag~TT Haug and Marius J{\o}hndal. 2008.
\newblock Creating a parallel treebank of the old {Indo-European} {B}ible
  translations.
\newblock In \emph{Proceedings of the second workshop on language technology
  for cultural heritage data (LaTeCH 2008)}, pages 27--34.

\bibitem[{Heinecke and Tyers(2019)}]{heinecke-tyers-2019-development}
Johannes Heinecke and Francis~M. Tyers. 2019.
\newblock \href {https://www.aclweb.org/anthology/W19-6904} {Development of a
  {U}niversal {D}ependencies treebank for {W}elsh}.
\newblock In \emph{Proceedings of the Celtic Language Technology Workshop},
  pages 21--31, Dublin, Ireland. European Association for Machine Translation.

\bibitem[{Hellwig et~al.(2020)Hellwig, Scarlata, Ackermann, and
  Widmer}]{hellwig-etal-2020-treebank}
Oliver Hellwig, Salvatore Scarlata, Elia Ackermann, and Paul Widmer. 2020.
\newblock \href {https://www.aclweb.org/anthology/2020.lrec-1.632} {The
  treebank of {V}edic {S}anskrit}.
\newblock In \emph{Proceedings of the 12th Language Resources and Evaluation
  Conference}, pages 5137--5146, Marseille, France. European Language Resources
  Association.

\bibitem[{Hladk{\'a} et~al.(2008)Hladk{\'a}, Haji{\v{c}}, Hana,
  Hlav{\'a}{\v{c}}ov{\'a}, M{\'\i}rovsk{\`y}, and Raab}]{hladka2008czech}
Barbora Hladk{\'a}, Jan Haji{\v{c}}, Jirka Hana, Jaroslava
  Hlav{\'a}{\v{c}}ov{\'a}, Ji{\v{r}}{\'\i} M{\'\i}rovsk{\`y}, and Jan Raab.
  2008.
\newblock The {C}zech academic corpus 2.0 guide.
\newblock \emph{The Prague Bulletin of Mathematical Linguistics}, 89(1):41--96.

\bibitem[{Howard and Ruder(2018)}]{howard-ruder-2018-universal}
Jeremy Howard and Sebastian Ruder. 2018.
\newblock \href {https://doi.org/10.18653/v1/P18-1031} {Universal language
  model fine-tuning for text classification}.
\newblock In \emph{Proceedings of the 56th Annual Meeting of the Association
  for Computational Linguistics (Volume 1: Long Papers)}, pages 328--339,
  Melbourne, Australia. Association for Computational Linguistics.

\bibitem[{Hu et~al.(2020)Hu, Ruder, Siddhant, Neubig, Firat, and
  Johnson}]{hu2020xtreme}
Junjie Hu, Sebastian Ruder, Aditya Siddhant, Graham Neubig, Orhan Firat, and
  Melvin Johnson. 2020.
\newblock \href {http://proceedings.mlr.press/v119/hu20b.html} {{XTREME}: A
  massively multilingual multi-task benchmark for evaluating cross-lingual
  generalization}.
\newblock In \emph{Proceedings of the 37th International Conference on Machine
  Learning}, volume 119, pages 4411--4421.

\bibitem[{Ingason et~al.(2020)Ingason, R{\"o}gnvaldsson, Sigurosson, and
  Wallenberg}]{20.500.12537/92}
Anton~Karl Ingason, Eir{\'{\i}}kur R{\"o}gnvaldsson, Einar~Freyr Sigurosson,
  and Joel~C. Wallenberg. 2020.
\newblock \href {http://hdl.handle.net/20.500.12537/92} {The {F}aroese parsed
  historical corpus}.
\newblock {CLARIN}-{IS}, Stofnun {\'A}rna Magn{\'u}ssonar.

\bibitem[{Ishola and Zeman(2020)}]{ishola-zeman-2020-yoruba}
Ol{\'a}j{\'\i}d{\'e} Ishola and Daniel Zeman. 2020.
\newblock \href {https://www.aclweb.org/anthology/2020.lrec-1.637}
  {{Y}or{\`u}b{\'a} dependency treebank ({YTB})}.
\newblock In \emph{Proceedings of the 12th Language Resources and Evaluation
  Conference}, pages 5178--5186, Marseille, France. European Language Resources
  Association.

\bibitem[{Jel{\'\i}nek(2017)}]{jelinek2017fictree}
Tom{\'a}s Jel{\'\i}nek. 2017.
\newblock Fic{T}ree: {A} manually annotated treebank of {C}zech fiction.
\newblock In \emph{ITAT}, pages 181--185.

\bibitem[{Johannsen et~al.(2015)Johannsen, Alonso, and
  Plank}]{johannsen2015universal}
Anders Johannsen, H{\'e}ctor~Mart{\'\i}nez Alonso, and Barbara Plank. 2015.
\newblock Universal dependencies for {D}anish.
\newblock In \emph{International Workshop on Treebanks and Linguistic Theories
  (TLT14)}, page 157.

\bibitem[{J{\'o}nsd{\'o}ttir and
  Ingason(2020)}]{jonsdottir-ingason-2020-creating}
Hildur J{\'o}nsd{\'o}ttir and Anton~Karl Ingason. 2020.
\newblock \href {https://www.aclweb.org/anthology/2020.lrec-1.357} {Creating a
  parallel {I}celandic dependency treebank from raw text to {U}niversal
  {D}ependencies}.
\newblock In \emph{Proceedings of the 12th Language Resources and Evaluation
  Conference}, pages 2924--2931, Marseille, France. European Language Resources
  Association.

\bibitem[{Kanerva(2020)}]{ud_finnish_ood_2020}
Jenna Kanerva. 2020.
\newblock {UD\_Finnish-OOD}.
\newblock \url{https://github.com/UniversalDependencies/UD_Finnish-OOD}.

\bibitem[{Kondratyuk and Straka(2019)}]{kondratyuk-straka-2019-75}
Dan Kondratyuk and Milan Straka. 2019.
\newblock \href {https://doi.org/10.18653/v1/D19-1279} {75 languages, 1 model:
  Parsing universal dependencies universally}.
\newblock In \emph{Proceedings of the 2019 Conference on Empirical Methods in
  Natural Language Processing and the 9th International Joint Conference on
  Natural Language Processing (EMNLP-IJCNLP)}, pages 2779--2795, Hong Kong,
  China. Association for Computational Linguistics.

\bibitem[{Kopacewicz(2018)}]{ud_akkadian_pisandub_2018}
Kamil Kopacewicz. 2018.
\newblock {UD\_Akkadian-PISANDUB}.
\newblock \url{https://github.com/UniversalDependencies/UD_Akkadian-PISANDUB}.

\bibitem[{Kotsyba et~al.(2018)Kotsyba, Moskalevskyi, Romanenko, Samoridna,
  Kosovska, Lytvyn, Orlenko, Brovko, Matushko, Onyshchuk, Pareviazko, Rychyk,
  Stetsenko, Umanets, and Masenko}]{UD_Ukrainian-IU}
Natalia Kotsyba, Bohdan Moskalevskyi, Mykhailo Romanenko, Halyna Samoridna,
  Ivanka Kosovska, Olha Lytvyn, Oksana Orlenko, Hanna Brovko, Bohdana Matushko,
  Natalia Onyshchuk, Valeriia Pareviazko, Yaroslava Rychyk, Anastasiia
  Stetsenko, Snizhana Umanets, and Larysa Masenko. 2018.
\newblock {UD\_Ukrainian-IU}.
\newblock \url{https://github.com/UniversalDependencies/UD_Ukrainian-IU}.

\bibitem[{Kr{\'\i}{\v{z}} et~al.(2016)Kr{\'\i}{\v{z}}, Hladk{\'a}, and
  Uresova}]{krivz2016czech}
Vincent Kr{\'\i}{\v{z}}, Barbora Hladk{\'a}, and Zdenka Uresova. 2016.
\newblock Czech legal text treebank 1.0.
\newblock In \emph{Proceedings of the Tenth International Conference on
  Language Resources and Evaluation (LREC'16)}, pages 2387--2392.

\bibitem[{Lacheret-Dujour et~al.(2019)Lacheret-Dujour, Kahane, and
  Pietrandrea}]{lacheret2019rhapsodie}
Anne Lacheret-Dujour, Sylvain Kahane, and Paola Pietrandrea. 2019.
\newblock \emph{Rhapsodie: A prosodic and syntactic treebank for spoken
  {F}rench}, volume~89.
\newblock John Benjamins Publishing Company.

\bibitem[{Lafferty et~al.(2001)Lafferty, McCallum, and
  Pereira}]{lafferty2001crf}
John~D. Lafferty, Andrew McCallum, and Fernando C.~N. Pereira. 2001.
\newblock \href
  {https://repository.upenn.edu/cgi/viewcontent.cgi?article=1162&context=cis_papers}
  {Conditional random fields: Probabilistic models for segmenting and labeling
  sequence data}.
\newblock In \emph{Proceedings of the Eighteenth International Conference on
  Machine Learning}, ICML '01, page 282–289, San Francisco, CA, USA.

\bibitem[{Lazaridou et~al.(2015)Lazaridou, Pham, and
  Baroni}]{lazaridou-etal-2015-combining}
Angeliki Lazaridou, Nghia~The Pham, and Marco Baroni. 2015.
\newblock \href {https://doi.org/10.3115/v1/N15-1016} {Combining language and
  vision with a multimodal skip-gram model}.
\newblock In \emph{Proceedings of the 2015 Conference of the North {A}merican
  Chapter of the Association for Computational Linguistics: Human Language
  Technologies}, pages 153--163, Denver, Colorado. Association for
  Computational Linguistics.

\bibitem[{Lee et~al.(2017)Lee, Leung, and Li}]{lee-etal-2017-towards}
John Lee, Herman Leung, and Keying Li. 2017.
\newblock \href {https://www.aclweb.org/anthology/W17-0408} {Towards
  {U}niversal {D}ependencies for learner {C}hinese}.
\newblock In \emph{Proceedings of the {N}o{D}a{L}i{D}a 2017 Workshop on
  Universal Dependencies ({UDW} 2017)}, pages 67--71, Gothenburg, Sweden.
  Association for Computational Linguistics.

\bibitem[{Levesque et~al.(2012)Levesque, Davis, and
  Morgenstern}]{levesque2012winograd}
Hector Levesque, Ernest Davis, and Leora Morgenstern. 2012.
\newblock \href {http://www.aaai.org/ocs/index.php/KR/KR12/paper/view/4492}
  {The {W}inograd schema challenge}.
\newblock In \emph{Thirteenth International Conference on the Principles of
  Knowledge Representation and Reasoning}, Rome, Italy.

\bibitem[{Liu et~al.(2020)Liu, Duh, Liu, and Gao}]{liu2020deep}
Xiaodong Liu, Kevin Duh, Liyuan Liu, and Jianfeng Gao. 2020.
\newblock \href {https://arxiv.org/abs/2008.07772v2} {Very deep transformers
  for neural machine translation}.
\newblock \emph{arXiv preprint arXiv:2008.07772v2}.

\bibitem[{Liu et~al.(2018)Liu, Zhu, Che, Qin, Schneider, and
  Smith}]{liu-etal-2018-parsing}
Yijia Liu, Yi~Zhu, Wanxiang Che, Bing Qin, Nathan Schneider, and Noah~A. Smith.
  2018.
\newblock \href {https://doi.org/10.18653/v1/N18-1088} {Parsing tweets into
  {U}niversal {D}ependencies}.
\newblock In \emph{Proceedings of the 2018 Conference of the North {A}merican
  Chapter of the Association for Computational Linguistics: Human Language
  Technologies, Volume 1 (Long Papers)}, pages 965--975, New Orleans,
  Louisiana. Association for Computational Linguistics.

\bibitem[{Liu et~al.(2019)Liu, Ott, Goyal, Du, Joshi, Chen, Levy, Lewis,
  Zettlemoyer, and Stoyanov}]{liu2019roberta}
Yinhan Liu, Myle Ott, Naman Goyal, Jingfei Du, Mandar Joshi, Danqi Chen, Omer
  Levy, Mike Lewis, Luke Zettlemoyer, and Veselin Stoyanov. 2019.
\newblock \href {https://arxiv.org/abs/1907.11692} {{RoBERTa}: A robustly
  optimized {BERT} pretraining approach}.
\newblock \emph{arXiv preprint arXiv:1907.11692}.

\bibitem[{Luukko et~al.(2020)Luukko, Sahala, Hardwick, and
  Lind{\'e}n}]{luukko-etal-2020-akkadian}
Mikko Luukko, Aleksi Sahala, Sam Hardwick, and Krister Lind{\'e}n. 2020.
\newblock \href {https://doi.org/10.18653/v1/2020.tlt-1.11} {{A}kkadian
  treebank for early neo-assyrian royal inscriptions}.
\newblock In \emph{Proceedings of the 19th International Workshop on Treebanks
  and Linguistic Theories}, pages 124--134, D{\"u}sseldorf, Germany.
  Association for Computational Linguistics.

\bibitem[{Lyashevskaya(2019)}]{lyashevskaya2019reusable}
Olga Lyashevskaya. 2019.
\newblock \href {http://www.dialog-21.ru/media/4614/lyashevskayaon-163.pdf} {A
  reusable tagset for the morphologically rich language in change: A case of
  {Middle Russian}}.
\newblock In \emph{Proceedings of the International Conference Dialogue 2019},
  pages 422--434.

\bibitem[{Lyashevskaya et~al.(2017)Lyashevskaya, Peljak-Łapińska, and
  Petrova}]{ud_belarusian_hse_2017}
Olga Lyashevskaya, Angelika Peljak-Łapińska, and Daria Petrova. 2017.
\newblock {UD\_Belarusian-HSE}.
\newblock \url{https://github.com/UniversalDependencies/UD_Belarusian-HSE}.

\bibitem[{Lyashevskaya and Sichinava(2017)}]{ud_lithuanian_hse_2017}
Olga Lyashevskaya and Dmitry Sichinava. 2017.
\newblock {UD\_Lithuanian-HSE}.
\newblock \url{https://github.com/UniversalDependencies/UD_Lithuanian-HSE}.

\bibitem[{Lynn and Foster(2016)}]{lynn2016universal}
Teresa Lynn and Jennifer Foster. 2016.
\newblock Universal dependencies for irish.
\newblock In \emph{CLTW}.

\bibitem[{Makazhanov et~al.(2015)Makazhanov, Sultangazina, Makhambetov, and
  Yessenbayev}]{makazhan_tl2015}
Aibek Makazhanov, Aitolkyn Sultangazina, Olzhas Makhambetov, and Zhandos
  Yessenbayev. 2015.
\newblock Syntactic annotation of {K}azakh: Following the {Universal
  Dependencies} guidelines. a report.
\newblock In \emph{3rd International Conference on Turkic Languages Processing,
  (TurkLang 2015)}, pages 338--350.

\bibitem[{Manning(2015)}]{manning2015computational}
Christopher~D Manning. 2015.
\newblock \href {https://doi.org/10.1162/COLI_a_00239} {Computational
  linguistics and deep learning}.
\newblock \emph{Computational Linguistics}, 41(4):701--707.

\bibitem[{M{\u{a}}r{\u{a}}nduc et~al.(2016)M{\u{a}}r{\u{a}}nduc, Perez, and
  Simionescu}]{muaruanduc2016social}
C{\u{a}}t{\u{a}}lina M{\u{a}}r{\u{a}}nduc, Cenel-Augusto Perez, and Radu
  Simionescu. 2016.
\newblock Social media-processing {R}omanian chat and discourse analysis.
\newblock \emph{Computaci{\'o}n y Sistemas}, 20(3):405--414.

\bibitem[{Mart{\'\i}nez~Alonso and
  Plank(2017)}]{martinez-alonso-plank-2017-multitask}
H{\'e}ctor Mart{\'\i}nez~Alonso and Barbara Plank. 2017.
\newblock \href {https://www.aclweb.org/anthology/E17-1005} {When is multitask
  learning effective? semantic sequence prediction under varying data
  conditions}.
\newblock In \emph{Proceedings of the 15th Conference of the {E}uropean Chapter
  of the Association for Computational Linguistics: Volume 1, Long Papers},
  Valencia, Spain. Association for Computational Linguistics.

\bibitem[{Mart{\'\i}nez~Alonso et~al.(2016)Mart{\'\i}nez~Alonso, Seddah, and
  Sagot}]{martinez-alonso-etal-2016-noisy}
H{\'e}ctor Mart{\'\i}nez~Alonso, Djam{\'e} Seddah, and Beno{\^\i}t Sagot. 2016.
\newblock \href {https://www.aclweb.org/anthology/W16-3905} {From noisy
  questions to {M}inecraft texts: Annotation challenges in extreme syntax
  scenario}.
\newblock In \emph{Proceedings of the 2nd Workshop on Noisy User-generated Text
  ({WNUT})}, pages 13--23, Osaka, Japan. The COLING 2016 Organizing Committee.

\bibitem[{McDonald et~al.(2013)McDonald, Nivre, Quirmbach-Brundage, Goldberg,
  Das, Ganchev, Hall, Petrov, Zhang, T{\"a}ckstr{\"o}m, Bedini,
  Bertomeu~Castell{\'o}, and Lee}]{mcdonald-etal-2013-universal}
Ryan McDonald, Joakim Nivre, Yvonne Quirmbach-Brundage, Yoav Goldberg, Dipanjan
  Das, Kuzman Ganchev, Keith Hall, Slav Petrov, Hao Zhang, Oscar
  T{\"a}ckstr{\"o}m, Claudia Bedini, N{\'u}ria Bertomeu~Castell{\'o}, and
  Jungmee Lee. 2013.
\newblock \href {https://www.aclweb.org/anthology/P13-2017} {{U}niversal
  {D}ependency annotation for multilingual parsing}.
\newblock In \emph{Proceedings of the 51st Annual Meeting of the Association
  for Computational Linguistics (Volume 2: Short Papers)}, pages 92--97, Sofia,
  Bulgaria. Association for Computational Linguistics.

\bibitem[{Mitrofan et~al.(2019)Mitrofan, Barbu~Mititelu, and
  Mitrofan}]{mitrofan-etal-2019-monero}
Maria Mitrofan, Verginica Barbu~Mititelu, and Grigorina Mitrofan. 2019.
\newblock \href {https://doi.org/10.18653/v1/W19-5008} {{M}o{NER}o: a
  biomedical gold standard corpus for the {R}omanian language}.
\newblock In \emph{Proceedings of the 18th BioNLP Workshop and Shared Task},
  pages 71--79, Florence, Italy. Association for Computational Linguistics.

\bibitem[{Mojiri et~al.(2020)Mojiri, Amir, Aghaei, and Ahmadi}]{ud_Soi-AHA2020}
Foroushani Mojiri, Hossein Amir, Hamid Aghaei, and Amir Ahmadi. 2020.
\newblock {UD\_Soi-AHA}.
\newblock \url{https://github.com/UniversalDependencies/UD_Soi-AHA}.

\bibitem[{Mojiri~Foroushani et~al.(2020{\natexlab{a}})Mojiri~Foroushani,
  Aghaei, and Ahmadi}]{ud_khunsari_aha_2020}
AmirHossein Mojiri~Foroushani, Hamid Aghaei, and Amir Ahmadi.
  2020{\natexlab{a}}.
\newblock {UD\_Khunsari-AHA}.
\newblock \url{https://github.com/UniversalDependencies/UD_Khunsari-AHA}.

\bibitem[{Mojiri~Foroushani et~al.(2020{\natexlab{b}})Mojiri~Foroushani,
  Aghaei, and Ahmadi}]{ud_nayini_aha_2020}
AmirHossein Mojiri~Foroushani, Hamid Aghaei, and Amir Ahmadi.
  2020{\natexlab{b}}.
\newblock {UD\_Nayini-AHA}.
\newblock \url{https://github.com/UniversalDependencies/UD_Nayini-AHA}.

\bibitem[{Muischnek et~al.(2014)Muischnek, M{\"u}{\"u}risep, Puolakainen,
  Aedmaa, Kirt, and S{\"a}rg}]{muischnek2014estonian}
Kadri Muischnek, Kaili M{\"u}{\"u}risep, Tiina Puolakainen, Eleri Aedmaa, Riin
  Kirt, and Dage S{\"a}rg. 2014.
\newblock Estonian dependency treebank and its annotation scheme.
\newblock In \emph{Proceedings of 13th workshop on treebanks and linguistic
  theories (TLT13)}, pages 285--291.

\bibitem[{Muischnek et~al.(2019)Muischnek, M{\"u}{\"u}risep, and
  S{\"a}rg}]{muischnek2019cg}
Kadri Muischnek, Kaili M{\"u}{\"u}risep, and Dage~Dage S{\"a}rg. 2019.
\newblock {CG} roots of {UD} treebank of {E}stonian web language.
\newblock In \emph{Proceedings of the NoDaLiDa 2019 Workshop on Constraint
  Grammar-Methods, Tools and Applications, 30 September 2019, Turku, Finland},
  168, pages 23--26. Link{\"o}ping University Electronic Press.

\bibitem[{Munro(2020)}]{munro2020human}
Robert Munro. 2020.
\newblock Human-in-the-loop machine learning.
\newblock \emph{Sl: O’REILLY MEDIA}.

\bibitem[{Nguyen et~al.(2009)Nguyen, Vu, Nguyen, Nguyen, and
  Le}]{nguyen-etal-2009-building}
Phuong-Thai Nguyen, Xuan-Luong Vu, Thi-Minh-Huyen Nguyen, Van-Hiep Nguyen, and
  Hong-Phuong Le. 2009.
\newblock \href {https://www.aclweb.org/anthology/W09-3035} {Building a large
  syntactically-annotated corpus of {V}ietnamese}.
\newblock In \emph{Proceedings of the Third Linguistic Annotation Workshop
  ({LAW} {III})}, pages 182--185, Suntec, Singapore. Association for
  Computational Linguistics.

\bibitem[{van Noord et~al.(2020)van Noord, Toral, and
  Bos}]{van-noord-etal-2020-character}
Rik van Noord, Antonio Toral, and Johan Bos. 2020.
\newblock \href {https://doi.org/10.18653/v1/2020.emnlp-main.371}
  {Character-level representations improve {DRS}-based semantic parsing {E}ven
  in the age of {BERT}}.
\newblock In \emph{Proceedings of the 2020 Conference on Empirical Methods in
  Natural Language Processing (EMNLP)}, pages 4587--4603, Online. Association
  for Computational Linguistics.

\bibitem[{Ojha and Zeman(2020)}]{ojha-zeman-2020-universal}
Atul~Kr. Ojha and Daniel Zeman. 2020.
\newblock \href {https://www.aclweb.org/anthology/2020.wildre-1.7} {{U}niversal
  {D}ependency treebanks for low-resource {I}ndian languages: The case of
  {B}hojpuri}.
\newblock In \emph{Proceedings of the WILDRE5{--} 5th Workshop on Indian
  Language Data: Resources and Evaluation}, pages 33--38, Marseille, France.
  European Language Resources Association (ELRA).

\bibitem[{Omura et~al.(2017)Omura, Takahashi, and Asahara}]{omura2017universal}
Mai Omura, Yuta Takahashi, and Masayuki Asahara. 2017.
\newblock Universal dependency for modern {J}apanese.
\newblock In \emph{Proceedings of the 7th Conference of Japanese Association
  for Digital Humanities (JADH2017)}, pages 34--36.

\bibitem[{{\"O}stling et~al.(2017){\"O}stling, B{\"o}rstell, G{\"a}rdenfors,
  and Wir{\'e}n}]{ostling-etal-2017-universal}
Robert {\"O}stling, Carl B{\"o}rstell, Moa G{\"a}rdenfors, and Mats Wir{\'e}n.
  2017.
\newblock \href {https://www.aclweb.org/anthology/W17-0243} {{U}niversal
  {D}ependencies for {S}wedish {S}ign {L}anguage}.
\newblock In \emph{Proceedings of the 21st Nordic Conference on Computational
  Linguistics}, pages 303--308, Gothenburg, Sweden. Association for
  Computational Linguistics.

\bibitem[{{\O}vrelid and Hohle(2016)}]{ovrelid-hohle-2016-universal}
Lilja {\O}vrelid and Petter Hohle. 2016.
\newblock \href {https://www.aclweb.org/anthology/L16-1250} {{U}niversal
  {D}ependencies for {N}orwegian}.
\newblock In \emph{Proceedings of the Tenth International Conference on
  Language Resources and Evaluation ({LREC}'16)}, pages 1579--1585,
  Portoro{\v{z}}, Slovenia. European Language Resources Association (ELRA).

\bibitem[{{\O}vrelid et~al.(2018){\O}vrelid, K{\aa}sen, Hagen, N{\o}klestad,
  Solberg, and Johannessen}]{ovrelid-etal-2018-lia}
Lilja {\O}vrelid, Andre K{\aa}sen, Kristin Hagen, Anders N{\o}klestad, Per~Erik
  Solberg, and Janne~Bondi Johannessen. 2018.
\newblock \href {https://www.aclweb.org/anthology/L18-1710} {The {LIA} treebank
  of spoken {N}orwegian dialects}.
\newblock In \emph{Proceedings of the Eleventh International Conference on
  Language Resources and Evaluation ({LREC} 2018)}, Miyazaki, Japan. European
  Language Resources Association (ELRA).

\bibitem[{Palmer et~al.(2009)Palmer, Bhatt, Narasimhan, Rambow, Sharma, and
  Xia}]{palmer2009hindi}
Martha Palmer, Rajesh Bhatt, Bhuvana Narasimhan, Owen Rambow, Dipti~Misra
  Sharma, and Fei Xia. 2009.
\newblock Hindi syntax: Annotating dependency, lexical predicate-argument
  structure, and phrase structure.
\newblock In \emph{The 7th International Conference on Natural Language
  Processing}, pages 14--17.

\bibitem[{Partanen et~al.(2018)Partanen, Blokland, Lim, Poibeau, and
  Rie{\ss}ler}]{partanen-etal-2018-first}
Niko Partanen, Rogier Blokland, KyungTae Lim, Thierry Poibeau, and Michael
  Rie{\ss}ler. 2018.
\newblock \href {https://doi.org/10.18653/v1/W18-6015} {The first
  {K}omi-{Z}yrian {U}niversal {D}ependencies treebanks}.
\newblock In \emph{Proceedings of the Second Workshop on Universal Dependencies
  ({UDW} 2018)}, pages 126--132, Brussels, Belgium. Association for
  Computational Linguistics.

\bibitem[{Paszke et~al.(2019)Paszke, Gross, Massa, Lerer, Bradbury, Chanan,
  Killeen, Lin, Gimelshein, Antiga, Desmaison, Kopf, Yang, DeVito, Raison,
  Tejani, Chilamkurthy, Steiner, Fang, Bai, and Chintala}]{pytorch}
Adam Paszke, Sam Gross, Francisco Massa, Adam Lerer, James Bradbury, Gregory
  Chanan, Trevor Killeen, Zeming Lin, Natalia Gimelshein, Luca Antiga, Alban
  Desmaison, Andreas Kopf, Edward Yang, Zachary DeVito, Martin Raison, Alykhan
  Tejani, Sasank Chilamkurthy, Benoit Steiner, Lu~Fang, Junjie Bai, and Soumith
  Chintala. 2019.
\newblock \href
  {http://papers.nips.cc/paper/9015-pytorch-an-imperative-style-high-performance-deep-learning-library.pdf}
  {{PyTorch}: An imperative style, high-performance deep learning library}.
\newblock In \emph{Advances in Neural Information Processing Systems 32}, pages
  8026--8037. Vancouver, Canada.

\bibitem[{Patejuk and Przepiórkowski(2018)}]{pat:prz:18:book}
Agnieszka Patejuk and Adam Przepiórkowski. 2018.
\newblock \href {http://nlp.ipipan.waw.pl/Bib/pat:prz:18:book.pdf} {\emph{From
  {L}exical {F}unctional {G}rammar to Enhanced {U}niversal {D}ependencies:
  Linguistically informed treebanks of {P}olish}}.
\newblock Institute of Computer Science, Polish Academy of Sciences, Warsaw.

\bibitem[{Peters et~al.(2018)Peters, Neumann, Iyyer, Gardner, Clark, Lee, and
  Zettlemoyer}]{peters-etal-2018-deep}
Matthew Peters, Mark Neumann, Mohit Iyyer, Matt Gardner, Christopher Clark,
  Kenton Lee, and Luke Zettlemoyer. 2018.
\newblock \href {https://doi.org/10.18653/v1/N18-1202} {Deep contextualized
  word representations}.
\newblock In \emph{Proceedings of the 2018 Conference of the North {A}merican
  Chapter of the Association for Computational Linguistics: Human Language
  Technologies, Volume 1 (Long Papers)}, pages 2227--2237, New Orleans,
  Louisiana. Association for Computational Linguistics.

\bibitem[{Piitulainen and Nurmi(2017)}]{ud_finnish_ftb_2017}
Jussi Piitulainen and Hanna Nurmi. 2017.
\newblock {UD\_Finnish-FTB}.
\newblock \url{https://github.com/UniversalDependencies/UD_Finnish-FTB}.

\bibitem[{Pirinen(2019)}]{pirinen-2019-building}
Tommi~A Pirinen. 2019.
\newblock \href {https://doi.org/10.18653/v1/W19-8016} {Building minority
  dependency treebanks, dictionaries and computational grammars at the same
  time{---}an experiment in {K}arelian treebanking}.
\newblock In \emph{Proceedings of the Third Workshop on Universal Dependencies
  (UDW, SyntaxFest 2019)}, pages 132--136, Paris, France. Association for
  Computational Linguistics.

\bibitem[{Plank et~al.(2016)Plank, S{\o}gaard, and Goldberg}]{plank-etal-2016}
Barbara Plank, Anders S{\o}gaard, and Yoav Goldberg. 2016.
\newblock \href {https://doi.org/10.18653/v1/P16-2067} {Multilingual
  part-of-speech tagging with bidirectional long short-term memory models and
  auxiliary loss}.
\newblock In \emph{Proceedings of the 54th Annual Meeting of the Association
  for Computational Linguistics (Volume 2: Short Papers)}, pages 412--418,
  Berlin, Germany. Association for Computational Linguistics.

\bibitem[{Prokopidis and
  Papageorgiou(2017)}]{prokopidis-papageorgiou-2017-universal}
Prokopis Prokopidis and Haris Papageorgiou. 2017.
\newblock \href {https://www.aclweb.org/anthology/W17-0413} {{U}niversal
  {D}ependencies for {G}reek}.
\newblock In \emph{Proceedings of the {N}o{D}a{L}i{D}a 2017 Workshop on
  Universal Dependencies ({UDW} 2017)}, pages 102--106, Gothenburg, Sweden.
  Association for Computational Linguistics.

\bibitem[{Pruksachatkun et~al.(2020)Pruksachatkun, Yeres, Liu, Phang, Htut,
  Wang, Tenney, and Bowman}]{pruksachatkun-etal-2020-jiant}
Yada Pruksachatkun, Phil Yeres, Haokun Liu, Jason Phang, Phu~Mon Htut, Alex
  Wang, Ian Tenney, and Samuel~R. Bowman. 2020.
\newblock \href {https://doi.org/10.18653/v1/2020.acl-demos.15} {jiant: A
  software toolkit for research on general-purpose text understanding models}.
\newblock In \emph{Proceedings of the 58th Annual Meeting of the Association
  for Computational Linguistics: System Demonstrations}, pages 109--117,
  Online. Association for Computational Linguistics.

\bibitem[{Pyysalo et~al.(2015)Pyysalo, Kanerva, Missil{\"a}, Laippala, and
  Ginter}]{pyysalo-etal-2015-universal}
Sampo Pyysalo, Jenna Kanerva, Anna Missil{\"a}, Veronika Laippala, and Filip
  Ginter. 2015.
\newblock \href {https://www.aclweb.org/anthology/W15-1821} {{U}niversal
  {D}ependencies for {F}innish}.
\newblock In \emph{Proceedings of the 20th Nordic Conference of Computational
  Linguistics ({NODALIDA} 2015)}, pages 163--172, Vilnius, Lithuania.
  Link{\"o}ping University Electronic Press, Sweden.

\bibitem[{Qi and Yasuoka(2019)}]{ud_chinese_gsdsimp_2019}
Peng Qi and Koichi Yasuoka. 2019.
\newblock {UD\_Chinese-GSDSimp}.
\newblock \url{https://github.com/UniversalDependencies/UD_Chinese-GSDSimp}.

\bibitem[{Rademaker et~al.(2017)Rademaker, Chalub, Real, Freitas, Bick, and
  de~Paiva}]{rademaker-etal-2017-universal}
Alexandre Rademaker, Fabricio Chalub, Livy Real, Cl{\'a}udia Freitas, Eckhard
  Bick, and Valeria de~Paiva. 2017.
\newblock \href {https://www.aclweb.org/anthology/W17-6523} {{U}niversal
  {D}ependencies for {P}ortuguese}.
\newblock In \emph{Proceedings of the Fourth International Conference on
  Dependency Linguistics (Depling 2017)}, pages 197--206, Pisa,Italy.
  Link{\"o}ping University Electronic Press.

\bibitem[{Radford et~al.(2019)Radford, Wu, Child, Luan, Amodei, and
  Sutskever}]{radford2019language}
Alec Radford, Jeffrey Wu, Rewon Child, David Luan, Dario Amodei, and Ilya
  Sutskever. 2019.
\newblock \href
  {https://cdn.openai.com/better-language-models/language_models_are_unsupervised_multitask_learners.pdf}
  {Language models are unsupervised multitask learners}.
\newblock \emph{OpenAI blog}.

\bibitem[{Rajpurkar et~al.(2018)Rajpurkar, Jia, and
  Liang}]{rajpurkar-etal-2018-know}
Pranav Rajpurkar, Robin Jia, and Percy Liang. 2018.
\newblock \href {https://doi.org/10.18653/v1/P18-2124} {Know what you don{'}t
  know: Unanswerable questions for {SQ}u{AD}}.
\newblock In \emph{Proceedings of the 56th Annual Meeting of the Association
  for Computational Linguistics (Volume 2: Short Papers)}, pages 784--789,
  Melbourne, Australia. Association for Computational Linguistics.

\bibitem[{Rama and Vajjala(2017)}]{rama-vajjala-2017-telugu}
Taraka Rama and Sowmya Vajjala. 2017.
\newblock \href {https://www.aclweb.org/anthology/W17-7616} {A {T}elugu
  treebank based on a grammar book}.
\newblock In \emph{Proceedings of the 16th International Workshop on Treebanks
  and Linguistic Theories}, pages 119--128, Prague, Czech Republic.

\bibitem[{Ramasamy and \v{Z}abokrtsk\'{y}(2012)}]{ta}
Loganathan Ramasamy and Zden\v{e}k \v{Z}abokrtsk\'{y}. 2012.
\newblock \href
  {http://www.lrec-conf.org/proceedings/lrec2012/summaries/456.html} {Prague
  dependency style treebank for {Tamil}}.
\newblock In \emph{Proceedings of Eighth International Conference on Language
  Resources and Evaluation ({LREC} 2012)}, pages 1888--1894, \.{I}stanbul,
  Turkey.

\bibitem[{Ravishankar(2017)}]{ravishankar-2017-universal}
Vinit Ravishankar. 2017.
\newblock \href {https://www.aclweb.org/anthology/W17-7623} {A {U}niversal
  {D}ependencies treebank for {M}arathi}.
\newblock In \emph{Proceedings of the 16th International Workshop on Treebanks
  and Linguistic Theories}, pages 190--200, Prague, Czech Republic.

\bibitem[{Rehbein et~al.(2019)Rehbein, Ruppenhofer, and
  Do}]{rehbein-etal-2019-tweede}
Ines Rehbein, Josef Ruppenhofer, and Bich-Ngoc Do. 2019.
\newblock \href {https://doi.org/10.18653/v1/W19-7811} {twee{D}e {--} a
  {U}niversal {D}ependencies treebank for {G}erman tweets}.
\newblock In \emph{Proceedings of the 18th International Workshop on Treebanks
  and Linguistic Theories (TLT, SyntaxFest 2019)}, pages 100--108, Paris,
  France. Association for Computational Linguistics.

\bibitem[{R{\"o}gnvaldsson et~al.(2012)R{\"o}gnvaldsson, Ingason, Sigurosson,
  and Wallenberg}]{rognvaldsson-etal-2012-icelandic}
Eir{\'\i}kur R{\"o}gnvaldsson, Anton~Karl Ingason, Einar~Freyr Sigurosson, and
  Joel Wallenberg. 2012.
\newblock \href
  {http://www.lrec-conf.org/proceedings/lrec2012/pdf/440_Paper.pdf} {The
  {I}celandic parsed historical corpus ({I}ce{P}a{HC})}.
\newblock In \emph{Proceedings of the Eighth International Conference on
  Language Resources and Evaluation ({LREC}'12)}, pages 1977--1984, Istanbul,
  Turkey. European Language Resources Association (ELRA).

\bibitem[{Ruder(2017)}]{ruder2017overview}
Sebastian Ruder. 2017.
\newblock \href {https://arxiv.org/abs/1706.05098} {An overview of multi-task
  learning in deep neural networks}.
\newblock \emph{arXiv preprint arXiv:1706.05098}.

\bibitem[{Ruder and Plank(2018)}]{ruder-plank-2018-strong}
Sebastian Ruder and Barbara Plank. 2018.
\newblock \href {https://doi.org/10.18653/v1/P18-1096} {Strong baselines for
  neural semi-supervised learning under domain shift}.
\newblock In \emph{Proceedings of the 56th Annual Meeting of the Association
  for Computational Linguistics (Volume 1: Long Papers)}, pages 1044--1054,
  Melbourne, Australia. Association for Computational Linguistics.

\bibitem[{Rueter and Partanen(2019)}]{rueter2019survey}
Jack Rueter and Niko Partanen. 2019.
\newblock Survey of {U}ralic {U}niversal {D}ependencies development.
\newblock In \emph{Workshop on Universal Dependencies}, page~78. The
  Association for Computational Linguistics.

\bibitem[{Rueter et~al.(2020)Rueter, Partanen, and
  Ponomareva}]{rueter-etal-2020-questions}
Jack Rueter, Niko Partanen, and Larisa Ponomareva. 2020.
\newblock \href {https://www.aclweb.org/anthology/2020.iwclul-1.3} {On the
  questions in developing computational infrastructure for {K}omi-permyak}.
\newblock In \emph{Proceedings of the Sixth International Workshop on
  Computational Linguistics of Uralic Languages}, pages 15--25, Wien, Austria.
  Association for Computational Linguistics.

\bibitem[{Rueter and Tyers(2018)}]{rueter2018towards}
Jack Rueter and Francis Tyers. 2018.
\newblock Towards an open-source universal-dependency treebank for {E}rzya.
\newblock In \emph{Proceedings of the Fourth International Workshop on
  Computational Linguistics of Uralic Languages}, pages 106--118.

\bibitem[{Rueter(2018)}]{rueter2018mordva}
Jack~Michael Rueter. 2018.
\newblock Mordva.
\newblock In \emph{The Uralic Languages}. Routledge.

\bibitem[{Sadegh~Rasooli et~al.(2020)Sadegh~Rasooli, Safari, Moloodi, and
  Nourian}]{sadegh2020persian}
Mohammad Sadegh~Rasooli, Pegah Safari, Amirsaeid Moloodi, and Alireza Nourian.
  2020.
\newblock The {P}ersian dependency treebank made universal.
\newblock \emph{arXiv e-prints}, pages arXiv--2009.

\bibitem[{Salomoni(2019)}]{LIT}
Alessio Salomoni. 2019.
\newblock {UD\_German-LIT}.
\newblock \url{https://github.com/UniversalDependencies/UD_German-LIT}.

\bibitem[{Samard{\v{z}}i{\'c} et~al.(2017)Samard{\v{z}}i{\'c}, Starovi{\'c},
  Agi{\'c}, and Ljube{\v{s}}i{\'c}}]{samardzic-etal-2017-universal}
Tanja Samard{\v{z}}i{\'c}, Mirjana Starovi{\'c}, {\v{Z}}eljko Agi{\'c}, and
  Nikola Ljube{\v{s}}i{\'c}. 2017.
\newblock \href {https://doi.org/10.18653/v1/W17-1407} {{U}niversal
  {D}ependencies for {S}erbian in comparison with {C}roatian and other {S}lavic
  languages}.
\newblock In \emph{Proceedings of the 6th Workshop on {B}alto-{S}lavic Natural
  Language Processing}, pages 39--44, Valencia, Spain. Association for
  Computational Linguistics.

\bibitem[{Samson and C{\"o}ltekin(2020)}]{UD_Tagalog-TRG}
Stephanie Samson and Cagr{\i} C{\"o}ltekin. 2020.
\newblock {UD\_Tagalog-TRG}.
\newblock \url{https://github.com/UniversalDependencies/UD_Tagalog-TRG}.

\bibitem[{Sanguinetti and Bosco(2014)}]{Sanguinetti2014}
Manuela Sanguinetti and Cristina Bosco. 2014.
\newblock Towards a {U}niversal {S}tanford {D}ependencies parallel treebank.
\newblock In \emph{Proceedings of the 13th Workshop on Treebanks and Linguistic
  Theories (TLT-13)}. Springer.

\bibitem[{Sanguinetti et~al.(2018)Sanguinetti, Bosco, Lavelli, Mazzei,
  Antonelli, and Tamburini}]{sanguinetti-etal-2018-postwita}
Manuela Sanguinetti, Cristina Bosco, Alberto Lavelli, Alessandro Mazzei, Oronzo
  Antonelli, and Fabio Tamburini. 2018.
\newblock \href {https://www.aclweb.org/anthology/L18-1279} {{P}o{STWITA}-{UD}:
  an {I}talian {T}witter treebank in {U}niversal {D}ependencies}.
\newblock In \emph{Proceedings of the Eleventh International Conference on
  Language Resources and Evaluation ({LREC} 2018)}, Miyazaki, Japan. European
  Language Resources Association (ELRA).

\bibitem[{Sanh et~al.(2019)Sanh, Wolf, and Ruder}]{sanh2019hierarchical}
Victor Sanh, Thomas Wolf, and Sebastian Ruder. 2019.
\newblock \href {https://ojs.aaai.org/index.php/AAAI/article/view/4673} {A
  hierarchical multi-task approach for learning embeddings from semantic
  tasks}.
\newblock In \emph{Proceedings of the AAAI Conference on Artificial
  Intelligence}, volume~33, pages 6949--6956, Honolulu, Hawaii, USA.

\bibitem[{Sarveswaran and Dias(2020)}]{sarveswaran2020thamizhiudp}
Kengatharaiyer Sarveswaran and Gihan Dias. 2020.
\newblock {ThamizhiUDp}: A dependency parser for {T}amil.
\newblock \emph{arXiv preprint arXiv:2012.13436}.

\bibitem[{Scannell(2020)}]{scannell-2020-universal}
Kevin Scannell. 2020.
\newblock \href {https://www.aclweb.org/anthology/2020.udw-1.17} {{U}niversal
  {D}ependencies for {M}anx {G}aelic}.
\newblock In \emph{Proceedings of the Fourth Workshop on Universal Dependencies
  (UDW 2020)}, pages 152--157, Barcelona, Spain (Online). Association for
  Computational Linguistics.

\bibitem[{Seddah and Candito(2016)}]{seddah-candito-2016-hard}
Djam{\'e} Seddah and Marie Candito. 2016.
\newblock \href {https://www.aclweb.org/anthology/L16-1375} {Hard time parsing
  questions: Building a {Q}uestion{B}ank for {F}rench}.
\newblock In \emph{Proceedings of the Tenth International Conference on
  Language Resources and Evaluation ({LREC}'16)}, pages 2366--2370,
  Portoro{\v{z}}, Slovenia. European Language Resources Association (ELRA).

\bibitem[{Seraji et~al.(2016)Seraji, Ginter, and
  Nivre}]{seraji-etal-2016-universal}
Mojgan Seraji, Filip Ginter, and Joakim Nivre. 2016.
\newblock \href {https://www.aclweb.org/anthology/L16-1374} {{U}niversal
  {D}ependencies for {P}ersian}.
\newblock In \emph{Proceedings of the Tenth International Conference on
  Language Resources and Evaluation ({LREC}'16)}, pages 2361--2365,
  Portoro{\v{z}}, Slovenia. European Language Resources Association (ELRA).

\bibitem[{Seyoum et~al.(2018)Seyoum, Miyao, and
  Mekonnen}]{seyoum-etal-2018-universal}
Binyam~Ephrem Seyoum, Yusuke Miyao, and Baye~Yimam Mekonnen. 2018.
\newblock \href {https://www.aclweb.org/anthology/L18-1350} {{U}niversal
  {D}ependencies for {A}mharic}.
\newblock In \emph{Proceedings of the Eleventh International Conference on
  Language Resources and Evaluation ({LREC} 2018)}, Miyazaki, Japan. European
  Language Resources Association (ELRA).

\bibitem[{Shavrina and Shapovalova(2017)}]{shavrina2017methodology}
Tatiana Shavrina and Olga Shapovalova. 2017.
\newblock To the methodology of corpus construction for machine
  learning:“{T}aiga” syntax tree corpus and parser.
\newblock In \emph{Proceedings of “CORPORA-2017” International Conference},
  pages 78--84.

\bibitem[{Shen et~al.(2016)Shen, McDonald, Zeman, and Qi}]{ud_chinese_gsd_2016}
Mo~Shen, Ryan McDonald, Daniel Zeman, and Peng Qi. 2016.
\newblock {UD\_Chinese-GSD}.
\newblock \url{https://github.com/UniversalDependencies/UD_Chinese-GSD}.

\bibitem[{Shopen(2018)}]{Warlpiri}
Timothy Shopen. 2018.
\newblock {UD\_Warlpiri-UFAL}.
\newblock \url{https://github.com/UniversalDependencies/UD_Lithuanian-HSE}.

\bibitem[{Silveira et~al.(2014)Silveira, Dozat, de~Marneffe, Bowman, Connor,
  Bauer, and Manning}]{silveira-etal-2014-gold}
Natalia Silveira, Timothy Dozat, Marie-Catherine de~Marneffe, Samuel Bowman,
  Miriam Connor, John Bauer, and Chris Manning. 2014.
\newblock \href
  {http://www.lrec-conf.org/proceedings/lrec2014/pdf/1089_Paper.pdf} {A gold
  standard dependency corpus for {E}nglish}.
\newblock In \emph{Proceedings of the Ninth International Conference on
  Language Resources and Evaluation ({LREC}'14)}, pages 2897--2904, Reykjavik,
  Iceland. European Language Resources Association (ELRA).

\bibitem[{Simov et~al.(2005)Simov, Osenova, Simov, and
  Kouylekov}]{SimOsSimKo2005}
Kiril Simov, Petya Osenova, Alexander Simov, and Milen Kouylekov. 2005.
\newblock Design and implementation of the {Bulgarian HPSG}-based treebank.
\newblock \emph{Journal of Research on Language and Computation. Special
  Issue}, pages 495--522.

\bibitem[{Socher et~al.(2013)Socher, Perelygin, Wu, Chuang, Manning, Ng, and
  Potts}]{socher-etal-2013-recursive}
Richard Socher, Alex Perelygin, Jean Wu, Jason Chuang, Christopher~D. Manning,
  Andrew Ng, and Christopher Potts. 2013.
\newblock \href {https://www.aclweb.org/anthology/D13-1170} {Recursive deep
  models for semantic compositionality over a sentiment treebank}.
\newblock In \emph{Proceedings of the 2013 Conference on Empirical Methods in
  Natural Language Processing}, pages 1631--1642, Seattle, Washington, USA.
  Association for Computational Linguistics.

\bibitem[{S{\o}gaard and Goldberg(2016)}]{sogaard-goldberg-2016-deep}
Anders S{\o}gaard and Yoav Goldberg. 2016.
\newblock \href {https://doi.org/10.18653/v1/P16-2038} {Deep multi-task
  learning with low level tasks supervised at lower layers}.
\newblock In \emph{Proceedings of the 54th Annual Meeting of the Association
  for Computational Linguistics (Volume 2: Short Papers)}, pages 231--235,
  Berlin, Germany. Association for Computational Linguistics.

\bibitem[{Stein and Pr{\'e}vost(2013)}]{stein2013syntactic}
Achim Stein and Sophie Pr{\'e}vost. 2013.
\newblock Syntactic annotation of medieval texts.
\newblock \emph{New methods in historical corpora}, 3:275.

\bibitem[{Straka(2018)}]{straka-2018-udpipe}
Milan Straka. 2018.
\newblock \href {https://doi.org/10.18653/v1/K18-2020} {{UDP}ipe 2.0 prototype
  at {C}o{NLL} 2018 {UD} shared task}.
\newblock In \emph{Proceedings of the {C}o{NLL} 2018 Shared Task: Multilingual
  Parsing from Raw Text to Universal Dependencies}, pages 197--207, Brussels,
  Belgium. Association for Computational Linguistics.

\bibitem[{Stymne et~al.(2018)Stymne, de~Lhoneux, Smith, and
  Nivre}]{stymne-etal-2018-parser}
Sara Stymne, Miryam de~Lhoneux, Aaron Smith, and Joakim Nivre. 2018.
\newblock \href {https://doi.org/10.18653/v1/P18-2098} {Parser training with
  heterogeneous treebanks}.
\newblock In \emph{Proceedings of the 56th Annual Meeting of the Association
  for Computational Linguistics (Volume 2: Short Papers)}, pages 619--625,
  Melbourne, Australia. Association for Computational Linguistics.

\bibitem[{Sulubacak et~al.(2016)Sulubacak, Gokirmak, Tyers, {\c{C}}{\"o}ltekin,
  Nivre, and Eryi{\u{g}}it}]{sulubacak-etal-2016-universal}
Umut Sulubacak, Memduh Gokirmak, Francis Tyers, {\c{C}}a{\u{g}}r{\i}
  {\c{C}}{\"o}ltekin, Joakim Nivre, and G{\"u}l{\c{s}}en Eryi{\u{g}}it. 2016.
\newblock \href {https://www.aclweb.org/anthology/C16-1325} {{U}niversal
  {D}ependencies for {T}urkish}.
\newblock In \emph{Proceedings of {COLING} 2016, the 26th International
  Conference on Computational Linguistics: Technical Papers}, pages 3444--3454,
  Osaka, Japan. The COLING 2016 Organizing Committee.

\bibitem[{Sutskever et~al.(2014)Sutskever, Vinyals, and
  Le}]{sutskever2014sequence}
Ilya Sutskever, Oriol Vinyals, and Quoc~V Le. 2014.
\newblock \href
  {https://proceedings.neurips.cc/paper/2014/hash/a14ac55a4f27472c5d894ec1c3c743d2-Abstract.html}
  {Sequence to sequence learning with neural networks}.
\newblock In \emph{Advances in Neural Information Processing Systems},
  volume~27, pages 3104--3112, Montreal, Canada.

\bibitem[{Thomas(2019)}]{thomas-2019-universal}
Guillaume Thomas. 2019.
\newblock \href {https://doi.org/10.18653/v1/W19-8008} {{U}niversal
  {D}ependencies for {M}by{\'a} {G}uaran{\'\i}}.
\newblock In \emph{Proceedings of the Third Workshop on Universal Dependencies
  (UDW, SyntaxFest 2019)}, pages 70--77, Paris, France. Association for
  Computational Linguistics.

\bibitem[{Tjong Kim~Sang and
  De~Meulder(2003)}]{tjong-kim-sang-de-meulder-2003-introduction}
Erik~F. Tjong Kim~Sang and Fien De~Meulder. 2003.
\newblock \href {https://www.aclweb.org/anthology/W03-0419} {Introduction to
  the {C}o{NLL}-2003 shared task: Language-independent named entity
  recognition}.
\newblock In \emph{Proceedings of the Seventh Conference on Natural Language
  Learning at {HLT}-{NAACL} 2003}, pages 142--147.

\bibitem[{Toska et~al.(2020)Toska, Nivre, and
  Zeman}]{toska-etal-2020-universal}
Marsida Toska, Joakim Nivre, and Daniel Zeman. 2020.
\newblock \href {https://www.aclweb.org/anthology/2020.udw-1.20} {{U}niversal
  {D}ependencies for {A}lbanian}.
\newblock In \emph{Proceedings of the Fourth Workshop on Universal Dependencies
  (UDW 2020)}, pages 178--188, Barcelona, Spain (Online). Association for
  Computational Linguistics.

\bibitem[{T\"{u}rk et~al.(2020)T\"{u}rk, Atmaca, \c{S}aziye
  Bet\"{u}l~\"{O}zate\c{s}, Berk, Bedir, K\"{o}ksal, \"{O}zt\"{u}rk
  Ba\c{s}aran, G\"{u}ng\"{o}r, and \"{O}zg\"{u}r}]{trk2020resources}
Utku T\"{u}rk, Furkan Atmaca, \c{S}aziye Bet\"{u}l~\"{O}zate\c{s}, G\"{o}zde
  Berk, Seyyit~Talha Bedir, Abdullatif K\"{o}ksal, Balkiz \"{O}zt\"{u}rk
  Ba\c{s}aran, Tunga G\"{u}ng\"{o}r, and Arzucan \"{O}zg\"{u}r. 2020.
\newblock \href {http://arxiv.org/abs/2002.10416} {Resources for {T}urkish
  dependency parsing: Introducing the {BOUN} treebank and the {BoAT} annotation
  tool}.

\bibitem[{Tyers and Mishchenkova(2020)}]{tyers-mishchenkova-2020-dependency}
Francis Tyers and Karina Mishchenkova. 2020.
\newblock \href {https://www.aclweb.org/anthology/2020.udw-1.22} {Dependency
  annotation of noun incorporation in polysynthetic languages}.
\newblock In \emph{Proceedings of the Fourth Workshop on Universal Dependencies
  (UDW 2020)}, pages 195--204, Barcelona, Spain (Online). Association for
  Computational Linguistics.

\bibitem[{Tyers et~al.(2018)Tyers, Sheyanova, Martynova, Stepachev, and
  Vinogorodskiy}]{tyers-etal-2018-multi}
Francis Tyers, Mariya Sheyanova, Aleksandra Martynova, Pavel Stepachev, and
  Konstantin Vinogorodskiy. 2018.
\newblock \href {https://doi.org/10.18653/v1/W18-6017} {Multi-source synthetic
  treebank creation for improved cross-lingual dependency parsing}.
\newblock In \emph{Proceedings of the Second Workshop on Universal Dependencies
  ({UDW} 2018)}, pages 144--150, Brussels, Belgium. Association for
  Computational Linguistics.

\bibitem[{Tyers and Ravishankar(2018)}]{tyers-ravishankar-2018-prototype}
Francis~M Tyers and Vinit Ravishankar. 2018.
\newblock \href {https://www.aclweb.org/anthology/2018.jeptalnrecital-court.1}
  {A prototype dependency treebank for {B}reton}.
\newblock In \emph{Actes de la Conf{\'e}rence TALN. Volume 1 - Articles longs,
  articles courts de TALN}, pages 197--204, Rennes, France. ATALA.

\bibitem[{Tyers and Sheyanova(2017)}]{tyers-sheyanova-2017-annotation}
Francis~M. Tyers and Mariya Sheyanova. 2017.
\newblock \href {https://doi.org/10.18653/v1/W17-0607} {Annotation schemes in
  {N}orth {S}{\'a}mi dependency parsing}.
\newblock In \emph{Proceedings of the Third Workshop on Computational
  Linguistics for Uralic Languages}, pages 66--75, St. Petersburg, Russia.
  Association for Computational Linguistics.

\bibitem[{Vincze et~al.(2010)Vincze, Szauter, Alm{\'a}si, M{\'o}ra, Alexin, and
  Csirik}]{vincze-etal-2010-hungarian}
Veronika Vincze, D{\'o}ra Szauter, Attila Alm{\'a}si, Gy{\"o}rgy M{\'o}ra,
  Zolt{\'a}n Alexin, and J{\'a}nos Csirik. 2010.
\newblock \href
  {http://www.lrec-conf.org/proceedings/lrec2010/pdf/465_Paper.pdf}
  {{H}ungarian dependency treebank}.
\newblock In \emph{Proceedings of the Seventh International Conference on
  Language Resources and Evaluation ({LREC}'10)}, Valletta, Malta. European
  Language Resources Association (ELRA).

\bibitem[{Vydrin(2013)}]{vydrin2013bamana}
Valentin Vydrin. 2013.
\newblock {B}amana {R}eference {C}orpus ({BRC}).
\newblock \emph{Procedia-Social and Behavioral Sciences}, 95:75--80.

\bibitem[{Wagner et~al.(2020)Wagner, Barry, and
  Foster}]{wagner-etal-2020-treebank}
Joachim Wagner, James Barry, and Jennifer Foster. 2020.
\newblock \href {https://doi.org/10.18653/v1/2020.acl-main.778} {Treebank
  embedding vectors for out-of-domain dependency parsing}.
\newblock In \emph{Proceedings of the 58th Annual Meeting of the Association
  for Computational Linguistics}, pages 8812--8818, Online. Association for
  Computational Linguistics.

\bibitem[{Wang et~al.(2018)Wang, Singh, Michael, Hill, Levy, and
  Bowman}]{wang-etal-2018-glue}
Alex Wang, Amanpreet Singh, Julian Michael, Felix Hill, Omer Levy, and Samuel
  Bowman. 2018.
\newblock \href {https://doi.org/10.18653/v1/W18-5446} {{GLUE}: A multi-task
  benchmark and analysis platform for natural language understanding}.
\newblock In \emph{Proceedings of the 2018 {EMNLP} Workshop {B}lackbox{NLP}:
  Analyzing and Interpreting Neural Networks for {NLP}}, pages 353--355,
  Brussels, Belgium. Association for Computational Linguistics.

\bibitem[{Wang et~al.(2019)Wang, Yang, and Zhang}]{wang2019genesis}
Hongmin Wang, Jie Yang, and Yue Zhang. 2019.
\newblock From genesis to creole language: Transfer learning for {S}inglish
  {U}niversal {D}ependencies parsing and pos tagging.
\newblock \emph{ACM Transactions on Asian and Low-Resource Language Information
  Processing (TALLIP)}, 19(1):1--29.

\bibitem[{Warstadt et~al.(2019)Warstadt, Singh, and
  Bowman}]{warstadt2019neural}
Alex Warstadt, Amanpreet Singh, and Samuel~R Bowman. 2019.
\newblock \href {https://www.aclweb.org/anthology/Q19-1040/} {Neural network
  acceptability judgments}.
\newblock \emph{Transactions of the Association for Computational Linguistics},
  7:625--641.

\bibitem[{Williams et~al.(2018)Williams, Nangia, and Bowman}]{N18-1101}
Adina Williams, Nikita Nangia, and Samuel Bowman. 2018.
\newblock \href {http://aclweb.org/anthology/N18-1101} {A broad-coverage
  challenge corpus for sentence understanding through inference}.
\newblock In \emph{Proceedings of the 2018 Conference of the North American
  Chapter of the Association for Computational Linguistics: Human Language
  Technologies, Volume 1 (Long Papers)}, pages 1112--1122. Association for
  Computational Linguistics.

\bibitem[{Wolf et~al.(2020)Wolf, Debut, Sanh, Chaumond, Delangue, Moi, Cistac,
  Rault, Louf, Funtowicz, Davison, Shleifer, von Platen, Ma, Jernite, Plu, Xu,
  Le~Scao, Gugger, Drame, Lhoest, and Rush}]{Wolf2019HuggingFacesTS}
Thomas Wolf, Lysandre Debut, Victor Sanh, Julien Chaumond, Clement Delangue,
  Anthony Moi, Pierric Cistac, Tim Rault, Remi Louf, Morgan Funtowicz, Joe
  Davison, Sam Shleifer, Patrick von Platen, Clara Ma, Yacine Jernite, Julien
  Plu, Canwen Xu, Teven Le~Scao, Sylvain Gugger, Mariama Drame, Quentin Lhoest,
  and Alexander Rush. 2020.
\newblock \href {https://doi.org/10.18653/v1/2020.emnlp-demos.6} {Transformers:
  State-of-the-art natural language processing}.
\newblock In \emph{Proceedings of the 2020 Conference on Empirical Methods in
  Natural Language Processing: System Demonstrations}, pages 38--45, Online.
  Association for Computational Linguistics.

\bibitem[{Wong et~al.(2017)Wong, Gerdes, Leung, and
  Lee}]{wong-etal-2017-quantitative}
Tak-sum Wong, Kim Gerdes, Herman Leung, and John Lee. 2017.
\newblock \href {https://www.aclweb.org/anthology/W17-6530} {Quantitative
  comparative syntax on the {C}antonese-{M}andarin parallel dependency
  treebank}.
\newblock In \emph{Proceedings of the Fourth International Conference on
  Dependency Linguistics (Depling 2017)}, pages 266--275, Pisa,Italy.
  Link{\"o}ping University Electronic Press.

\bibitem[{Wr{\'o}blewska(2018)}]{wroblewska-2018-extended}
Alina Wr{\'o}blewska. 2018.
\newblock \href {https://doi.org/10.18653/v1/W18-6020} {Extended and enhanced
  {P}olish dependency bank in {U}niversal {D}ependencies format}.
\newblock In \emph{Proceedings of the Second Workshop on Universal Dependencies
  ({UDW} 2018)}, pages 173--182, Brussels, Belgium. Association for
  Computational Linguistics.

\bibitem[{Yako(2019)}]{ud_assyrian_as_2019}
Mary Yako. 2019.
\newblock {UD\_Assyrian-AS}.
\newblock \url{https://github.com/UniversalDependencies/UD_Assyrian-AS}.

\bibitem[{Yasuoka(2019)}]{yasuoka2019universal}
Koichi Yasuoka. 2019.
\newblock Universal dependencies treebank of the four books in {C}lassical
  {C}hinese.
\newblock In \emph{DADH2019: 10th International Conference of Digital Archives
  and Digital Humanities}, pages 20--28. Digital Archives and Digital
  Humanities.

\bibitem[{Yavrumyan et~al.(2017)Yavrumyan, Khachatrian, Danielyan, and
  Arakelyan}]{yavrumyan2017armtdp}
M~Yavrumyan, H~Khachatrian, A~Danielyan, and G~Arakelyan. 2017.
\newblock {ArmTDP: Eastern Armenian} treebank and dependency parser.
\newblock In \emph{XI International Conference on Armenian Linguistics,
  Abstracts. Yerevan}.

\bibitem[{Zaheer et~al.(2018)Zaheer, Reddi, Sachan, Kale, and
  Kumar}]{zaheer2018adaptive}
Manzil Zaheer, Sashank Reddi, Devendra Sachan, Satyen Kale, and Sanjiv Kumar.
  2018.
\newblock \href
  {https://papers.nips.cc/paper/2018/hash/90365351ccc7437a1309dc64e4db32a3-Abstract.html}
  {Adaptive methods for nonconvex optimization}.
\newblock In \emph{Advances in Neural Information Processing Systems},
  volume~31, pages 9793--9803, Montreal, Canada.

\bibitem[{Zahra(2020)}]{Madar}
Shorouq Zahra. 2020.
\newblock Parsing low-resource {Levantine Arabic}: Annotation projection versus
  small-sized annotated data.

\bibitem[{Zeldes(2017)}]{Zeldes2017}
Amir Zeldes. 2017.
\newblock \href {https://doi.org/http://dx.doi.org/10.1007/s10579-016-9343-x}
  {The {GUM} corpus: Creating multilayer resources in the classroom}.
\newblock \emph{Language Resources and Evaluation}, 51(3):581--612.

\bibitem[{Zeldes and Abrams(2018)}]{zeldes-abrams-2018-coptic}
Amir Zeldes and Mitchell Abrams. 2018.
\newblock \href {https://doi.org/10.18653/v1/W18-6022} {The {C}optic
  {U}niversal {D}ependency treebank}.
\newblock In \emph{Proceedings of the Second Workshop on Universal Dependencies
  ({UDW} 2018)}, pages 192--201, Brussels, Belgium. Association for
  Computational Linguistics.

\bibitem[{Zeman et~al.(2017)Zeman, Nedoluzhko, and
  Majli\v{s}}]{UD_Upper_Sorbian-UFAL}
Dan Zeman, Anna Nedoluzhko, and Martin Majli\v{s}. 2017.
\newblock {UD\_Upper\_Sorbian-UFAL}.
\newblock \url{https://github.com/UniversalDependencies/UD_Upper_Sorbian-UFAL}.

\bibitem[{Zeman(2017)}]{zeman2017slovak}
Daniel Zeman. 2017.
\newblock Slovak dependency treebank in {U}niversal {D}ependencies.
\newblock \emph{Journal of Linguistics/Jazykovedn{\`y} casopis},
  68(2):385--395.

\end{thebibliography}
\bibliographystyle{acl_natbib}

\appendix
\clearpage
\section*{Appendix}
\begin{table}
\centering
\begin{tabular}{l r r r r r r r r r}
\toprule
Dataset & RTE & MRPC & CoLa & SST-2 & QNLI & QQP & MNLI & MNLI-mis & SNLI \\
size & 2k & 4k & 9k & 67k & 105k & 364k & 393k & 393k & 550k \\
\midrule
Single & 67.1 & 85.5 & 74.7 & 88.4 & 85.2 & 90.5 & 80.2 & 80.8 & 88.9 \\
All & 69.3 & 81.6 & 70.2 & 88.2 & 82.3 & 90.1 & 79.2 & 79.7 & 88.1 \\
Smoothed & 72.9 & 82.8 & 72.7 & 87.6 & 83.1 & 90.3 & 78.8 & 80.1 & 88.4 \\
\bottomrule
\end{tabular}
\caption{The scores (accuracy) per dataset on the GLUE tasks (dev) for a
variety of multi-task settings (ordered by size, indicated in number of
sentences in training data).}
\label{tab:analysisGlue}
\end{table}

\paragraph{Multi-dataset evaluation on GLUE tasks} Table~\ref{tab:analysisGlue}
contains the per-dataset scores for the GLUE tasks for all our tested settings.
Only for RTE the performance increases when using multi-task learning. Overall,
smoothing helps to overcome some of the performance loss we get when training
one model on all datasets simultaneously.

\paragraph{Multi-dataset evaluation on UD treebanks} Table~\ref{tab:allUD} (on
the next four pages) shows the LAS scores for each treebank (UD2.7) for all of
our settings. We pre-processed the data with the UD-conversion tools to remove
all language-specific sub-labels, and the multi-word tokens and empty nodes.
\textit{However, we calculate the scores against the official files for fair
comparison}.\footnote{This is why the scores for some datasets might seem low
compared to previous work, which did either do tokenization or did not take it
into account during evaluation. In our case the model is punished for not
tokenizing.} We included as many datasets as we could find. In the top part of
the table, we include all official UD datasets for which we could get the words
(only UD\_Arabic-NYUAD and UD\_Japanese-BCCWJ are missing), and the last 12
treebanks are taken from other sources, some have undergone some specific
pre-processing to pass the evaluation script; details about this process can be
found in the repository in \texttt{scripts/udExtras}.

\begin{table*}
 \resizebox{\textwidth}{!}{
\begin{tabular}{p{2.2cm} p{4cm} p{1.5cm} r r r r r r r}

\toprule
 &  &  &  &  &  & \multicolumn{3}{|c|}{+smoothing} \\
dataset & citation & proxy & size & self & conc. & conc. & sepDec & dataEmb\\
\midrule
af\_afribooms & {\small \cite{dirix-etal-2017-universal}} & --- & 33,894 & 86.7 & 85.9 & 86.6 & \textbf{87.0} & 85.9 \\
aii\_as & {\small \cite{ud_assyrian_as_2019}} & et\_ewt & 0 & \textbf{9.7} & 3.5 & 3.9 & 5.1 & 3.4 \\
ajp\_madar & {\small \cite{Madar}} & ar\_padt & 0 & 31.2 & \textbf{33.8} & 33.1 & 33.2 & 31.2 \\
akk\_pisandub & {\small \cite{ud_akkadian_pisandub_2018}} & et\_edt & 0 & 3.0 & 4.3 & \textbf{4.7} & 3.6 & 3.3 \\
akk\_riao & {\small \cite{luukko-etal-2020-akkadian}} & et\_edt & 0 & 4.0 & \textbf{8.2} & 7.6 & 7.3 & 8.1 \\
am\_att & {\small \cite{seyoum-etal-2018-universal}} & et\_ewt & 0 & \textbf{1.8} & 0.8 & 0.8 & 0.5 & 0.8 \\
apu\_ufpa & {\small \cite{freitas2017posse}} & fi\_ftb & 0 & 6.1 & 13.3 & 13.1 & 8.1 & \textbf{13.4} \\
aqz\_tudet & {\small \cite{aragon2018variaccoes}} & cs\_pdt & 0 & 6.7 & 9.6 & 9.6 & 9.2 & \textbf{14.7} \\
ar\_padt & {\small \cite{UD_Arabic-PADT}} & --- & 191,869 & 31.5 & 31.4 & 31.3 & 31.4 & \textbf{31.5} \\
ar\_pud & {\small \cite{mcdonald-etal-2013-universal}} & ar\_padt & 0 & 62.8 & 64.5 & 63.9 & 64.0 & \textbf{64.7} \\
be\_hse & {\small \cite{ud_belarusian_hse_2017}} & --- & 249,897 & 81.0 & 83.6 & 83.1 & 81.8 & \textbf{83.8} \\
bg\_btb & {\small \cite{SimOsSimKo2005}} & --- & 124,336 & 92.5 & 92.7 & 92.5 & 92.7 & \textbf{92.7} \\
bho\_bhtb & {\small \cite{ojha-zeman-2020-universal}} & hi\_hdtb & 0 & \textbf{37.7} & 36.1 & 36.2 & 36.5 & 36.3 \\
bm\_crb & {\small \cite{vydrin2013bamana}} & qhe\_hiencs & 0 & \textbf{8.8} & 6.5 & 6.1 & 6.7 & 6.3 \\
br\_keb & {\small \cite{tyers-ravishankar-2018-prototype}} & fr\_gsd & 0 & \textbf{54.9} & 32.0 & 31.3 & 33.2 & 32.4 \\
bxr\_bdt & {\small \cite{badmaeva:2017}} & --- & 153 & 11.6 & 23.9 & \textbf{29.0} & 21.5 & 24.0 \\
ca\_ancora & {\small \cite{alonso2016universal}} & --- & 416,659 & 92.1 & \textbf{92.2} & 91.8 & 91.9 & 92.2 \\
ckt\_hse & \resizebox{4cm}{!}{\cite{tyers-mishchenkova-2020-dependency}} & ru\_syntagrus & 0 & 8.1 & \textbf{15.3} & 15.3 & 13.7 & 14.5 \\
cop\_scriptorium & {\small \cite{zeldes-abrams-2018-coptic}} & --- & 12,926 & 0.8 & 0.8 & 0.7 & 0.7 & \textbf{0.9} \\
cs\_cac & {\small \cite{hladka2008czech}} & --- & 471,594 & 91.2 & \textbf{92.2} & 91.0 & 90.8 & 92.0 \\
cs\_cltt & {\small \cite{krivz2016czech}} & --- & 27,752 & 83.9 & 89.6 & 88.7 & 87.7 & \textbf{89.6} \\
cs\_fictree & {\small \cite{jelinek2017fictree}} & --- & 133,137 & 91.5 & 93.0 & 92.3 & 92.4 & \textbf{93.3} \\
cs\_pdt & {\small \cite{bejvcek2013prague}} & --- & 1,171,190 & 92.7 & 92.8 & 91.2 & 91.1 & \textbf{92.8} \\
cs\_pud & {\small \cite{mcdonald-etal-2013-universal}} & cs\_pdt & 0 & 87.7 & \textbf{88.4} & 88.0 & 88.1 & 88.2 \\
cu\_proiel & {\small \cite{haug2008creating}} & --- & 37,432 & 65.1 & 67.1 & \textbf{68.0} & 67.6 & 67.3 \\
cy\_ccg & {\small \cite{heinecke-tyers-2019-development}} & --- & 15,706 & 74.5 & 73.9 & \textbf{76.2} & 76.0 & 73.8 \\
da\_ddt & {\small \cite{johannsen2015universal}} & --- & 80,378 & 86.7 & 86.1 & 86.5 & \textbf{86.8} & 86.0 \\
de\_gsd & {\small \cite{brants2004tiger}} & --- & 259,194 & 81.7 & 79.9 & 81.5 & \textbf{82.0} & 80.8 \\
de\_hdt & {\small \cite{borges-volker-etal-2019-hdt}} & --- & 2,753,627 & \textbf{96.7} & 96.6 & 90.0 & 94.8 & 96.6 \\
de\_lit & {\small \cite{LIT}} & de\_hdt & 0 & 76.9 & 78.9 & \textbf{79.8} & 77.8 & 78.4 \\
de\_pud & {\small \cite{mcdonald-etal-2013-universal}} & de\_hdt & 0 & 78.5 & 81.2 & \textbf{82.3} & 78.8 & 80.6 \\
el\_gdt & \resizebox{4cm}{!}{\cite{prokopidis-papageorgiou-2017-universal}} & --- & 41,212 & 86.9 & 89.0 & 88.9 & 88.8 & \textbf{89.0} \\
en\_ewt & {\small \cite{silveira-etal-2014-gold}} & --- & 202,141 & \textbf{87.6} & 85.6 & 85.4 & 86.7 & 86.0 \\
en\_gum & {\small \cite{Zeldes2017}} & --- & 81,861 & \textbf{89.0} & 87.3 & 87.3 & 88.9 & 88.1 \\
en\_lines & {\small \cite{ahrenberg-2015-converting}} & --- & 57,372 & 87.4 & 86.8 & 86.9 & \textbf{88.0} & 87.2 \\
en\_partut & {\small \cite{Sanguinetti2014}} & --- & 43,477 & 89.7 & 89.3 & 89.5 & \textbf{90.7} & 89.8 \\
en\_pronouns & {\small \cite{munro2020human}} & en\_ewt & 0 & 81.8 & 85.5 & 86.8 & 84.9 & \textbf{87.2} \\
en\_pud & {\small \cite{mcdonald-etal-2013-universal}} & en\_ewt & 0 & 89.3 & 87.8 & 87.7 & \textbf{89.7} & 89.1 \\
es\_ancora & {\small \cite{alonso2016universal}} & --- & 443,086 & \textbf{90.8} & 89.0 & 88.7 & 90.5 & 90.4 \\
es\_gsd & {\small \cite{mcdonald-etal-2013-universal}} & --- & 375,149 & 85.6 & 81.7 & 81.6 & \textbf{85.8} & 85.0 \\
es\_pud & {\small \cite{mcdonald-etal-2013-universal}} & es\_gsd & 0 & 79.4 & 78.6 & 78.7 & \textbf{79.7} & 79.5 \\
et\_edt & {\small \cite{muischnek2014estonian}} & --- & 344,646 & 86.7 & 86.7 & 85.5 & 85.5 & \textbf{86.8} \\
et\_ewt & {\small \cite{muischnek2019cg}} & --- & 34,287 & 74.6 & 82.4 & 81.6 & 80.9 & \textbf{82.4} \\
eu\_bdt & {\small \cite{aranzabe2015automatic}} & --- & 72,974 & \textbf{83.3} & 82.3 & 82.4 & 82.4 & 82.3 \\
fa\_perdt & {\small \cite{sadegh2020persian}} & --- & 445,587 & \textbf{89.2} & 88.9 & 84.2 & 87.8 & 89.2 \\
fa\_seraji & {\small \cite{seraji-etal-2016-universal}} & --- & 119,945 & \textbf{87.2} & 81.8 & 84.8 & 86.9 & 86.4 \\
fi\_ftb & {\small \cite{ud_finnish_ftb_2017}} & --- & 127,359 & \textbf{89.1} & 80.4 & 80.1 & 88.6 & 88.8 \\
fi\_ood & {\small \cite{ud_finnish_ood_2020}} & fi\_tdt & 0 & 77.6 & 69.5 & 69.1 & 77.5 & \textbf{78.1} \\
fi\_pud & {\small \cite{mcdonald-etal-2013-universal}} & fi\_tdt & 0 & 90.4 & 86.6 & 86.0 & \textbf{90.5} & 90.4 \\
fi\_tdt & {\small \cite{pyysalo-etal-2015-universal}} & --- & 162,617 & 89.1 & 83.2 & 82.7 & \textbf{89.5} & 89.5 \\
\bottomrule
\end{tabular}}
\end{table*}

\begin{table*}
 \resizebox{\textwidth}{!}{
\begin{tabular}{p{2.2cm} p{4cm} p{1.5cm} r r r r r r r}

\toprule
 &  &  &  &  &  & \multicolumn{3}{|c|}{+smoothing} \\
dataset & citation & proxy & size & self & conc. & conc. & sepDec & dataEmb\\
\midrule
fo\_farpahc & {\small \cite{20.500.12537/92}} & --- & 23,089 & 80.9 & 87.0 & 86.5 & 85.4 & \textbf{87.1} \\
fo\_oft & {\small \cite{tyers-etal-2018-multi}} & fo\_farpahc & 0 & 49.8 & 62.1 & 62.2 & 61.6 & \textbf{62.7} \\
fr\_fqb & {\small \cite{seddah-candito-2016-hard}} & fr\_gsd & 0 & 84.9 & 84.6 & 84.6 & \textbf{85.2} & 85.2 \\
fr\_gsd & {\small \cite{guillaume2019conversion}} & --- & 344,975 & \textbf{88.6} & 86.0 & 85.3 & 88.5 & 88.2 \\
fr\_partut & {\small \cite{Sanguinetti2014}} & --- & 23,322 & 87.0 & 81.7 & 82.7 & \textbf{87.7} & 82.7 \\
fr\_pud & {\small \cite{mcdonald-etal-2013-universal}} & fr\_gsd & 0 & 85.3 & 83.9 & 84.1 & 85.4 & \textbf{85.5} \\
fr\_sequoia & {\small \cite{bonfante2018application}} & --- & 49,157 & 88.4 & 85.9 & 87.1 & \textbf{89.6} & 87.4 \\
fr\_spoken & {\small \cite{lacheret2019rhapsodie}} & --- & 14,921 & 77.5 & 81.9 & \textbf{83.1} & 82.3 & 81.8 \\
fro\_srcmf & {\small \cite{stein2013syntactic}} & --- & 136,020 & \textbf{88.5} & 87.6 & 87.3 & 87.4 & 87.6 \\
ga\_idt & {\small \cite{lynn2016universal}} & --- & 95,860 & 77.8 & \textbf{78.1} & 78.1 & 77.9 & 78.1 \\
gd\_arcosg & {\small \cite{batchelor-2019-universal}} & --- & 37,817 & 72.2 & 72.8 & \textbf{73.7} & 73.7 & 72.8 \\
gl\_ctg & {\small \cite{gomez2017recursos}} & --- & 71,928 & \textbf{66.3} & 65.6 & 65.4 & 66.0 & 65.5 \\
gl\_treegal & {\small \cite{garcia2016universal}} & --- & 14,158 & 65.9 & 56.7 & 63.5 & \textbf{68.4} & 58.5 \\
got\_proiel & {\small \cite{haug2008creating}} & --- & 35,024 & 75.4 & 79.0 & \textbf{79.7} & 77.8 & 78.9 \\
grc\_perseus & {\small \cite{bamman2011ancient}} & --- & 159,895 & 59.6 & 63.3 & 62.4 & 62.2 & \textbf{63.4} \\
grc\_proiel & {\small \cite{eckhoff2018proiel}} & --- & 187,033 & 71.7 & 74.8 & 74.0 & 73.3 & \textbf{74.9} \\
gsw\_uzh & {\small \cite{aepli2018parsing}} & de\_hdt & 0 & 27.8 & 36.7 & \textbf{37.1} & 35.1 & 36.9 \\
gun\_thomas & {\small \cite{thomas-2019-universal}} & it\_isdt & 0 & 7.7 & 10.5 & \textbf{11.1} & 9.2 & 10.9 \\
gv\_cadhan & {\small \cite{scannell-2020-universal}} & en\_singpar & 0 & 2.9 & 12.2 & \textbf{13.4} & 6.3 & 12.5 \\
he\_htb & {\small \cite{mcdonald-etal-2013-universal}} & --- & 98,348 & \textbf{36.3} & 36.0 & 35.9 & 36.1 & 36.2 \\
hi\_hdtb & {\small \cite{palmer2009hindi}} & --- & 281,057 & \textbf{92.0} & 91.8 & 91.6 & 91.8 & 91.9 \\
hi\_pud & {\small \cite{mcdonald-etal-2013-universal}} & hi\_hdtb & 0 & 59.6 & \textbf{59.8} & 59.5 & 59.6 & 59.7 \\
hr\_set & {\small \cite{agic-ljubesic-2015-universal}} & --- & 152,857 & 89.1 & 89.5 & 88.9 & 89.7 & \textbf{90.0} \\
hsb\_ufal & {\small \cite{UD_Upper_Sorbian-UFAL}} & --- & 460 & 10.5 & 59.8 & \textbf{65.9} & 60.1 & 59.8 \\
hu\_szeged & {\small \cite{vincze-etal-2010-hungarian}} & --- & 20,166 & 82.6 & 83.9 & \textbf{85.1} & 84.8 & 84.0 \\
hy\_armtdp & {\small \cite{yavrumyan2017armtdp}} & --- & 41,837 & 75.0 & 76.8 & \textbf{77.3} & 76.6 & 76.2 \\
id\_csui & {\small \cite{alfina2020tree}} & --- & 17,904 & 77.1 & 74.8 & 76.9 & \textbf{79.2} & 75.1 \\
id\_gsd & {\small \cite{mcdonald-etal-2013-universal}} & --- & 97,531 & \textbf{79.9} & 79.7 & 79.3 & 79.5 & 79.9 \\
id\_pud & {\small \cite{mcdonald-etal-2013-universal}} & id\_gsd & 0 & 59.6 & \textbf{63.1} & 62.9 & 61.0 & 63.1 \\
is\_icepahc & {\small \cite{rognvaldsson-etal-2012-icelandic}} & --- & 704,716 & \textbf{83.5} & 83.4 & 80.3 & 80.0 & 83.4 \\
is\_pud & {\small \cite{jonsdottir-ingason-2020-creating}} & is\_icepahc & 0 & 57.9 & \textbf{59.3} & 59.0 & 58.7 & 59.3 \\
it\_isdt & {\small \cite{bosco2014evalita}} & --- & 257,616 & 81.1 & 81.0 & 80.8 & 81.0 & \textbf{81.4} \\
it\_partut & {\small \cite{Sanguinetti2014}} & --- & 45,477 & 79.2 & 80.0 & 80.1 & \textbf{80.7} & 80.3 \\
it\_postwita & {\small \cite{sanguinetti-etal-2018-postwita}} & --- & 95,308 & 74.0 & \textbf{74.9} & 74.8 & 74.8 & 74.7 \\
it\_pud & {\small \cite{mcdonald-etal-2013-universal}} & it\_isdt & 0 & 80.1 & 80.3 & 80.3 & \textbf{80.6} & 80.6 \\
it\_twittiro & {\small \cite{cignarella-etal-2019-presenting}} & --- & 22,656 & 72.6 & \textbf{77.3} & 77.1 & 76.5 & 76.6 \\
it\_vit & {\small \cite{alfieri2016almost}} & --- & 208,795 & 78.6 & 78.0 & 77.6 & \textbf{78.9} & 78.8 \\
ja\_gsd & {\small \cite{asahara-etal-2018-universal}} & --- & 167,482 & \textbf{93.1} & 92.7 & 92.4 & 92.4 & 92.6 \\
ja\_modern & {\small \cite{omura2017universal}} & ja\_gsd & 0 & 51.8 & 52.9 & 53.8 & \textbf{53.8} & 52.9 \\
ja\_pud & {\small \cite{mcdonald-etal-2013-universal}} & ja\_gsd & 0 & 94.3 & \textbf{94.3} & 94.1 & 94.2 & 94.2 \\
kfm\_aha & {\small \cite{ud_khunsari_aha_2020}} & fa\_perdt & 0 & 17.6 & 16.7 & 18.5 & 18.9 & \textbf{22.1} \\
kk\_ktb & {\small \cite{makazhan_tl2015}} & --- & 511 & 21.6 & 56.7 & \textbf{59.1} & 53.0 & 56.5 \\
kmr\_mg & {\small \cite{gokirmak-tyers-2017-dependency}} & --- & 242 & 12.0 & 15.8 & \textbf{36.0} & 28.4 & 16.1 \\
ko\_gsd & {\small \cite{chun-etal-2018-building}} & --- & 56,687 & \textbf{85.6} & 73.7 & 77.7 & 85.0 & 82.5 \\
ko\_kaist & {\small \cite{chun-etal-2018-building}} & --- & 296,446 & \textbf{87.6} & 85.0 & 80.3 & 86.2 & 87.1 \\
ko\_pud & {\small \cite{mcdonald-etal-2013-universal}} & ko\_kaist & 0 & 47.7 & 46.1 & 43.6 & 48.2 & \textbf{48.9} \\
koi\_uh & {\small \cite{rueter-etal-2020-questions}} & ru\_syntagrus & 0 & 12.2 & 19.1 & \textbf{19.4} & 18.0 & 18.4 \\
kpv\_ikdp & {\small \cite{partanen-etal-2018-first}} & ru\_syntagrus & 0 & 19.5 & 22.1 & \textbf{22.2} & 21.1 & 21.8 \\
kpv\_lattice & {\small \cite{partanen-etal-2018-first}} & ru\_syntagrus & 0 & 8.2 & 11.3 & \textbf{11.7} & 10.5 & 11.6 \\
krl\_kkpp & {\small \cite{pirinen-2019-building}} & fi\_tdt & 0 & 45.9 & 42.1 & 44.9 & 46.0 & \textbf{46.4} \\
\bottomrule
\end{tabular}}
\end{table*}

\begin{table*}
 \resizebox{\textwidth}{!}{
\begin{tabular}{p{2.2cm} p{4cm} p{1.5cm} r r r r r r r}

\toprule
 &  &  &  &  &  & \multicolumn{3}{|c|}{+smoothing} \\
dataset & citation & proxy & size & self & conc. & conc. & sepDec & dataEmb\\
\midrule
la\_ittb & {\small \cite{cecchini-etal-2018-challenges}} & --- & 390,785 & 90.5 & \textbf{91.0} & 89.5 & 89.8 & 91.0 \\
la\_llct & {\small \cite{cecchini-etal-2018-challenges}} & --- & 194,143 & \textbf{94.6} & 94.6 & 94.2 & 94.5 & 94.5 \\
la\_perseus & {\small \cite{bamman2011ancient}} & --- & 18,184 & 63.3 & 68.4 & 69.1 & \textbf{69.4} & 68.3 \\
la\_proiel & {\small \cite{haug2008creating}} & --- & 172,133 & 79.9 & \textbf{81.6} & 80.1 & 80.1 & 81.6 \\
lt\_alksnis & {\small \cite{bielinskiene2016lithuanian}} & --- & 47,641 & 78.0 & 78.1 & \textbf{78.3} & 78.3 & 78.2 \\
lt\_hse & \resizebox{4cm}{!}{\cite{ud_lithuanian_hse_2017}} & --- & 3,210 & 47.8 & 63.7 & 64.2 & \textbf{68.5} & 64.3 \\
lv\_lvtb & {\small \cite{gruzitis-etal-2018-creation}} & --- & 167,594 & \textbf{86.8} & 86.6 & 86.3 & 86.2 & 86.8 \\
lzh\_kyoto & {\small \cite{yasuoka2019universal}} & --- & 185,211 & 79.7 & \textbf{79.8} & 75.9 & 75.6 & 79.7 \\
mdf\_jr & {\small \cite{rueter2018mordva}} & ru\_syntagrus & 0 & 16.8 & 17.7 & 17.5 & \textbf{18.2} & 17.8 \\
mr\_ufal & {\small \cite{ravishankar-2017-universal}} & --- & 2,730 & 50.3 & 65.9 & \textbf{67.1} & 64.6 & 64.6 \\
mt\_mudt & {\small \cite{vceplo2018constituent}} & --- & 22,880 & 75.5 & 76.2 & \textbf{78.9} & 78.1 & 76.2 \\
myu\_tudet & {\small \cite{Tudet}} & ro\_nonstandard & 0 & 16.1 & 15.4 & \textbf{17.4} & 14.0 & 14.4 \\
myv\_jr & {\small \cite{rueter2018towards}} & be\_hse & 0 & \textbf{20.1} & 18.9 & 19.1 & 18.6 & 18.6 \\
nl\_alpino & {\small \cite{bouma-van-noord-2017-increasing}} & --- & 185,883 & 90.9 & 91.4 & 91.4 & 91.1 & \textbf{91.5} \\
nl\_lassysmall & {\small \cite{bouma-van-noord-2017-increasing}} & --- & 75,080 & 89.4 & 91.0 & 91.0 & 90.7 & \textbf{91.2} \\
no\_bokmaal & {\small \cite{ovrelid-hohle-2016-universal}} & --- & 243,886 & 92.2 & \textbf{92.6} & 92.2 & 92.3 & 92.5 \\
no\_nynorsk & {\small \cite{ovrelid-hohle-2016-universal}} & --- & 245,330 & 91.8 & 92.1 & 92.0 & 91.9 & \textbf{92.2} \\
no\_nynorsklia & {\small \cite{ovrelid-etal-2018-lia}} & --- & 35,207 & 74.1 & 75.6 & \textbf{76.0} & 75.4 & 75.8 \\
nyq\_aha & {\small \cite{ud_nayini_aha_2020}} & fa\_perdt & 0 & 30.8 & 29.1 & 37.2 & 34.2 & \textbf{38.9} \\
olo\_kkpp & {\small \cite{pirinen-2019-building}} & --- & 144 & 8.4 & 40.4 & \textbf{44.7} & 26.3 & 43.1 \\
orv\_rnc & {\small \cite{lyashevskaya2019reusable}} & --- & 10,156 & 58.3 & 70.6 & \textbf{71.6} & 69.6 & 70.5 \\
orv\_torot & \resizebox{4cm}{!}{\cite{eckhoff2015linguistics}} & --- & 118,630 & 63.9 & 65.1 & 64.6 & 64.4 & \textbf{65.4} \\
otk\_tonqq & {\small \cite{ud_old_turkish_tonqq_2020}} & et\_ewt & 0 & 7.7 & 11.8 & 5.9 & \textbf{11.9} & 7.1 \\
pcm\_nsc & {\small \cite{caron-etal-2019-surface}} & --- & 111,843 & 90.0 & 90.2 & 89.9 & 89.5 & \textbf{90.2} \\
pl\_lfg & \resizebox{4cm}{!}{\cite{pat:prz:18:book}} & --- & 104,750 & 95.7 & 93.7 & 93.6 & 95.7 & \textbf{95.8} \\
pl\_pdb & {\small \cite{wroblewska-2018-extended}} & --- & 279,596 & 89.4 & 88.8 & 88.2 & 89.3 & \textbf{89.7} \\
pl\_pud & {\small \cite{wroblewska-2018-extended}} & pl\_pdb & 0 & 91.2 & 91.0 & 90.5 & 91.0 & \textbf{91.4} \\
pt\_bosque & {\small \cite{rademaker-etal-2017-universal}} & --- & 191,406 & \textbf{78.2} & 74.1 & 73.8 & 78.1 & 77.1 \\
pt\_gsd & {\small \cite{mcdonald-etal-2013-universal}} & --- & 238,714 & \textbf{83.0} & 80.8 & 80.6 & 82.7 & 82.7 \\
pt\_pud & {\small \cite{mcdonald-etal-2013-universal}} & pt\_gsd & 0 & 68.5 & \textbf{69.6} & 69.3 & 68.8 & 68.8 \\
qtd\_sagt & {\small \cite{cetinoglu-coltekin-2019-challenges}} & --- & 4,761 & 46.4 & 58.0 & \textbf{60.9} & 59.9 & 57.7 \\
ro\_nonstandard & {\small \cite{muaruanduc2016social}} & --- & 532,881 & 86.8 & 87.0 & 86.0 & 85.7 & \textbf{87.1} \\
ro\_rrt & {\small \cite{barbu2016romanian}} & --- & 185,113 & 88.3 & \textbf{88.6} & 88.3 & 88.2 & 88.5 \\
ro\_simonero & {\small \cite{mitrofan-etal-2019-monero}} & --- & 116,857 & \textbf{91.3} & 91.0 & 91.2 & 91.0 & 91.0 \\
ru\_gsd & {\small \cite{mcdonald-etal-2013-universal}} & --- & 74,906 & 87.4 & 88.9 & 89.2 & 89.2 & \textbf{89.7} \\
ru\_pud & {\small \cite{mcdonald-etal-2013-universal}} & ru\_syntagrus & 0 & 86.8 & 88.5 & \textbf{89.0} & 86.9 & 87.4 \\
ru\_syntagrus & {\small \cite{droganova2018data}} & --- & 870,479 & \textbf{93.7} & 93.0 & 88.9 & 92.0 & 93.5 \\
ru\_taiga & \resizebox{4cm}{!}{\cite{shavrina2017methodology}} & --- & 43,557 & 77.9 & 78.7 & 79.6 & \textbf{81.0} & 80.1 \\
sa\_ufal & {\small \cite{dwivedi2017universal}} & hi\_hdtb & 0 & 14.2 & 15.5 & 16.2 & 14.4 & \textbf{16.5} \\
sa\_vedic & {\small \cite{hellwig-etal-2020-treebank}} & --- & 17,445 & 54.9 & 57.9 & \textbf{60.0} & 57.5 & 57.8 \\
sk\_snk & {\small \cite{zeman2017slovak}} & --- & 80,575 & 92.3 & \textbf{94.3} & 93.7 & 93.1 & 94.2 \\
sl\_ssj & {\small \cite{dobrovoljc-etal-2017-universal}} & --- & 112,530 & \textbf{93.4} & 93.2 & 93.1 & 93.0 & 93.0 \\
sl\_sst & {\small \cite{dobrovoljc-nivre-2016-universal}} & --- & 19,473 & 69.4 & 73.6 & \textbf{74.7} & 73.9 & 73.5 \\
sme\_giella & {\small \cite{tyers-sheyanova-2017-annotation}} & --- & 16,835 & 61.3 & 65.3 & \textbf{68.5} & 64.5 & 65.5 \\
sms\_giellagas & {\small \cite{rueter2019survey}} & id\_gsd & 0 & 7.8 & \textbf{14.9} & 14.6 & 11.7 & 14.8 \\
soj\_aha & {\small \cite{ud_Soi-AHA2020}} & fa\_perdt & 0 & 27.9 & 37.6 & 27.3 & 32.1 & \textbf{39.4} \\
sq\_tsa & {\small \cite{toska-etal-2020-universal}} & ga\_idt & 0 & 52.1 & 62.8 & \textbf{64.0} & 51.2 & 62.6 \\
sr\_set & {\small \cite{samardzic-etal-2017-universal}} & --- & 74,259 & 91.9 & 91.4 & 91.9 & 92.4 & \textbf{92.5} \\
sv\_lines & {\small \cite{ahrenberg-2015-converting}} & --- & 55,451 & 86.5 & \textbf{88.3} & 88.1 & 88.2 & 88.2 \\
sv\_pud & {\small \cite{mcdonald-etal-2013-universal}} & sv\_lines & 0 & 83.8 & \textbf{86.9} & 86.9 & 85.8 & 86.7 \\
\bottomrule
\end{tabular}}
\end{table*}

\begin{table*}
 \resizebox{\textwidth}{!}{
\begin{tabular}{p{2.2cm} p{4cm} p{1.5cm} r r r r r r r}

\toprule
 &  &  &  &  &  & \multicolumn{3}{|c|}{+smoothing} \\
dataset & citation & proxy & size & self & conc. & conc. & sepDec & dataEmb\\
\midrule
sv\_talbanken & {\small \cite{mcdonald-etal-2013-universal}} & --- & 66,645 & 89.1 & 89.8 & 89.7 & \textbf{90.1} & 89.7 \\
swl\_sslc & {\small \cite{ostling-etal-2017-universal}} & --- & 644 & 26.2 & 26.1 & \textbf{37.7} & 29.4 & 26.8 \\
ta\_mwtt & {\small \cite{sarveswaran2020thamizhiudp}} & ta\_ttb & 0 & 65.4 & \textbf{70.0} & 66.1 & 67.1 & 69.9 \\
ta\_ttb & \resizebox{4cm}{!}{\cite{ta}} & --- & 5,734 & 40.8 & 44.7 & 44.7 & \textbf{44.9} & 44.3 \\
te\_mtg & {\small \cite{rama-vajjala-2017-telugu}} & --- & 5,082 & 82.8 & 84.2 & 84.5 & \textbf{85.7} & 84.7 \\
th\_pud & {\small \cite{mcdonald-etal-2013-universal}} & en\_ewt & 0 & \textbf{28.2} & 25.7 & 25.4 & 22.2 & 26.2 \\
tl\_trg & {\small \cite{UD_Tagalog-TRG}} & en\_singpar & 0 & \textbf{34.8} & 32.9 & 29.9 & 25.0 & 32.4 \\
tl\_ugnayan & {\small \cite{UD_Tagalog-Ugnayan}} & en\_singpar & 0 & \textbf{28.4} & 24.9 & 25.0 & 19.3 & 27.4 \\
tpn\_tudet & {\small \cite{UD_Tupinamba-TuDeT}} & cs\_pdt & 0 & \textbf{9.7} & 5.1 & 4.2 & 6.5 & 3.2 \\
tr\_boun & {\small \cite{trk2020resources}} & --- & 97,257 & 69.6 & 68.8 & 67.1 & 69.9 & \textbf{70.0} \\
tr\_gb & {\small \cite{coltekin2015grammar}} & tr\_boun & 0 & 66.3 & 64.8 & 64.1 & 66.1 & \textbf{66.6} \\
tr\_imst & {\small \cite{sulubacak-etal-2016-universal}} & --- & 36,822 & 62.5 & 59.1 & 61.2 & \textbf{64.2} & 63.8 \\
tr\_pud & {\small \cite{mcdonald-etal-2013-universal}} & tr\_boun & 0 & 61.4 & 60.7 & 59.3 & 61.2 & \textbf{61.6} \\
ug\_udt & {\small \cite{eli-etal-2016-universal}} & --- & 19,262 & 48.5 & \textbf{50.3} & 50.1 & 49.7 & 50.2 \\
uk\_iu & {\small \cite{UD_Ukrainian-IU}} & --- & 92,355 & 88.0 & 90.2 & 89.7 & 89.6 & \textbf{90.3} \\
ur\_udtb & {\small \cite{bhathindi}} & --- & 108,690 & 81.6 & 82.4 & 82.3 & 82.2 & \textbf{82.8} \\
vi\_vtb & {\small \cite{nguyen-etal-2009-building}} & --- & 20,285 & \textbf{66.1} & 65.3 & 65.3 & 65.7 & 65.4 \\
wbp\_ufal & {\small \cite{Warlpiri}} & id\_gsd & 0 & 5.5 & 6.8 & \textbf{8.7} & 7.6 & 8.0 \\
wo\_wtb & {\small \cite{dione-2019-developing}} & --- & 22,817 & 67.6 & 68.5 & \textbf{72.6} & 71.4 & 68.4 \\
yo\_ytb & {\small \cite{ishola-zeman-2020-yoruba}} & ga\_idt & 0 & 16.0 & 17.2 & 14.4 & 12.7 & \textbf{18.1} \\
yue\_hk & {\small \cite{wong-etal-2017-quantitative}} & zh\_gsd & 0 & 31.8 & 32.4 & 32.5 & 31.7 & \textbf{32.7} \\
zh\_cfl & {\small \cite{lee-etal-2017-towards}} & zh\_gsdsimp & 0 & 47.4 & \textbf{48.1} & 47.6 & 46.9 & 47.9 \\
zh\_gsd & {\small \cite{ud_chinese_gsd_2016}} & --- & 98,616 & \textbf{85.9} & 84.2 & 84.4 & 84.3 & 84.0 \\
zh\_gsdsimp & {\small \cite{ud_chinese_gsdsimp_2019}} & --- & 98,616 & \textbf{85.8} & 84.1 & 84.5 & 84.3 & 84.2 \\
zh\_hk & {\small \cite{wong-etal-2017-quantitative}} & zh\_gsd & 0 & 52.1 & \textbf{53.7} & 53.5 & 52.9 & 53.6 \\
zh\_pud & {\small \cite{mcdonald-etal-2013-universal}} & zh\_gsd & 0 & 62.1 & 62.2 & 62.0 & 61.7 & \textbf{62.3} \\
\midrule
de\_tweede & {\small \cite{rehbein-etal-2019-tweede}} & --- & 5,752 & 68.2 & 76.9 & 77.6 & \textbf{79.6} & 77.7 \\
en\_aae & {\small \cite{blodgett-etal-2018-twitter}} & en\_ewt & 0 & 51.5 & 55.1 & 55.9 & \textbf{56.5} & 56.1 \\
en\_convbank & {\small \cite{davidson-etal-2019-dependency}} & --- & 5,057 & 69.1 & 71.4 & 70.4 & 71.2 & \textbf{71.9} \\
en\_esl & {\small \cite{berzak-etal-2016-universal}} & --- & 78,541 & 92.0 & 91.4 & 91.3 & \textbf{92.1} & 91.7 \\
en\_gumreddit & {\small \cite{behzad-zeldes-2020-cross}} & --- & 10,831 & 75.9 & 84.9 & 84.8 & \textbf{86.5} & 85.5 \\
en\_monoise & \resizebox{4cm}{!}{\cite{van-der-goot-van-noord-2018-modeling}} & en\_ewt & 0 & 55.6 & 64.7 & 64.5 & 62.4 & \textbf{64.7} \\
en\_singpar & {\small \cite{wang2019genesis}} & --- & 27,368 & 80.3 & 79.0 & 78.5 & \textbf{82.2} & 79.4 \\
en\_tweebank2 & {\small \cite{liu-etal-2018-parsing}} & --- & 24,753 & 80.5 & 81.7 & 82.4 & \textbf{82.6} & 81.6 \\
fr\_extremeugc & {\small \cite{martinez-alonso-etal-2016-noisy}} & fr\_gsd & 0 & 56.2 & 55.7 & 56.6 & \textbf{58.0} & 54.4 \\
fr\_ftb & {\small \cite{abeille-etal-2000-building}} & --- & 442,228 & \textbf{83.1} & 82.2 & 81.6 & 82.9 & 82.8 \\
qfn\_fame & \resizebox{4cm}{!}{\cite{braggaar2021}} & nl\_alpino & 0 & \textbf{54.0} & 43.2 & 42.6 & 43.8 & 43.4 \\
qhe\_hiencs & {\small \cite{bhat-etal-2018-universal}} & --- & 20,203 & 62.8 & 62.4 & \textbf{65.5} & 64.0 & 62.0 \\
\bottomrule
\end{tabular}}
\caption{LAS scores from official conll2018 script on test splits of all UD datasets we could obtain, averaged over 3 random seeds. Size refers to number of sentences in the training split. Results for single dataset trained models, and our 4 multi-task strategies. The last 12 rows contain datasets that are either available without words on the official Universal Dependencies website or are not officialy submitted.}
\label{tab:allUD}
\end{table*}

\end{document}